\documentclass[10pt,twocolumn,letterpaper]{article}

\usepackage[pagenumbers]{cvpr}

\usepackage{cuted} 

\usepackage{currfile} 

\usepackage{caption} 
\usepackage[most]{tcolorbox}

\newtcolorbox{promptbox}[1]{
    enhanced,
    breakable,
    colback=white,
    colframe=gray,
    boxrule=0.6pt,
    arc=2mm,
    left=3mm,
    right=3mm,
    top=5mm,
    bottom=3mm,
    before skip=12pt,
    after skip=12pt,
    fontupper=\small,
    fonttitle=\small\sffamily\bfseries,
    coltitle=black,
    title={#1},
    attach boxed title to top center={yshift=-2mm},
    boxed title style={
        colback=gray!25,
        colframe=gray!25,
        boxrule=0pt,
        arc=1.5mm,
        left=2mm,
        right=2mm,
        top=1mm,
        bottom=1mm
    }
}

\definecolor{cvprblue}{rgb}{0.21,0.49,0.74}
\usepackage[pagebackref,breaklinks,colorlinks,allcolors=cvprblue]{hyperref}

\title{ReWorldBench: Evaluating Code-Based Reconstruction and Editing of Realistic
3D Scenes}

\author{%
Ruibing Wang\textsuperscript{1*},
Congyue Zhou\textsuperscript{2*},
Zhen Fang\textsuperscript{1\ensuremath{\dagger}},
Xinge Peng\textsuperscript{1},
Yongxuan Lv\textsuperscript{1},
Haodong Li\textsuperscript{3}\\
Guohui Zhang\textsuperscript{1},
Xiaoxiao Ma\textsuperscript{1},
Lin Chen\textsuperscript{1},
Jie Huang\textsuperscript{1},
Feng Zhao\textsuperscript{1\ensuremath{\ddagger}}\\[0.4em]
{\small\textsuperscript{1}University of Science and Technology of China}\\
{\small\textsuperscript{2}South China University of Technology}\\
{\small\textsuperscript{3}Department of Computer Science and Engineering, The Chinese University of Hong Kong}\\[0.3em]
{\small
\textsuperscript{*}Equal contribution\quad
\textsuperscript{\ensuremath{\dagger}}Project leader\quad
\textsuperscript{\ensuremath{\ddagger}}Corresponding author}%
}

\usepackage{graphicx}
\usepackage{booktabs}
\usepackage{xcolor}
\usepackage{colortbl}

\definecolor{scoreblue}{RGB}{232,242,255}
\definecolor{scoregreen}{RGB}{228,255,228}
\definecolor{scoreyellow}{RGB}{255,248,220}

\newcommand{\modelicon}[1]{%
    \raisebox{-0.18em}{%
        \includegraphics[
            width=1.1em,
            height=1.1em,
            keepaspectratio
        ]{fig/model_icon/#1.png}%
    }\hspace{0.4em}%
}
\usepackage{graphicx,xcolor,tcolorbox}

\usepackage{xcolor}
\usepackage{tcolorbox}

\tcbset{
    rubricexample/.style={
        colback=gray!8,
        colframe=gray!40,
        boxrule=0.4pt,
        arc=1mm,
        boxsep=0pt,
        left=2mm,
        right=2mm,
        top=1.5mm,
        bottom=1.5mm,
        before skip=6pt,
        after skip=6pt,
        fontupper=\small
    }
}
\usepackage{float}
\begin{document}
\maketitle

\begin{strip}
\vspace{-13mm}
\centering
\includegraphics[width=\textwidth]{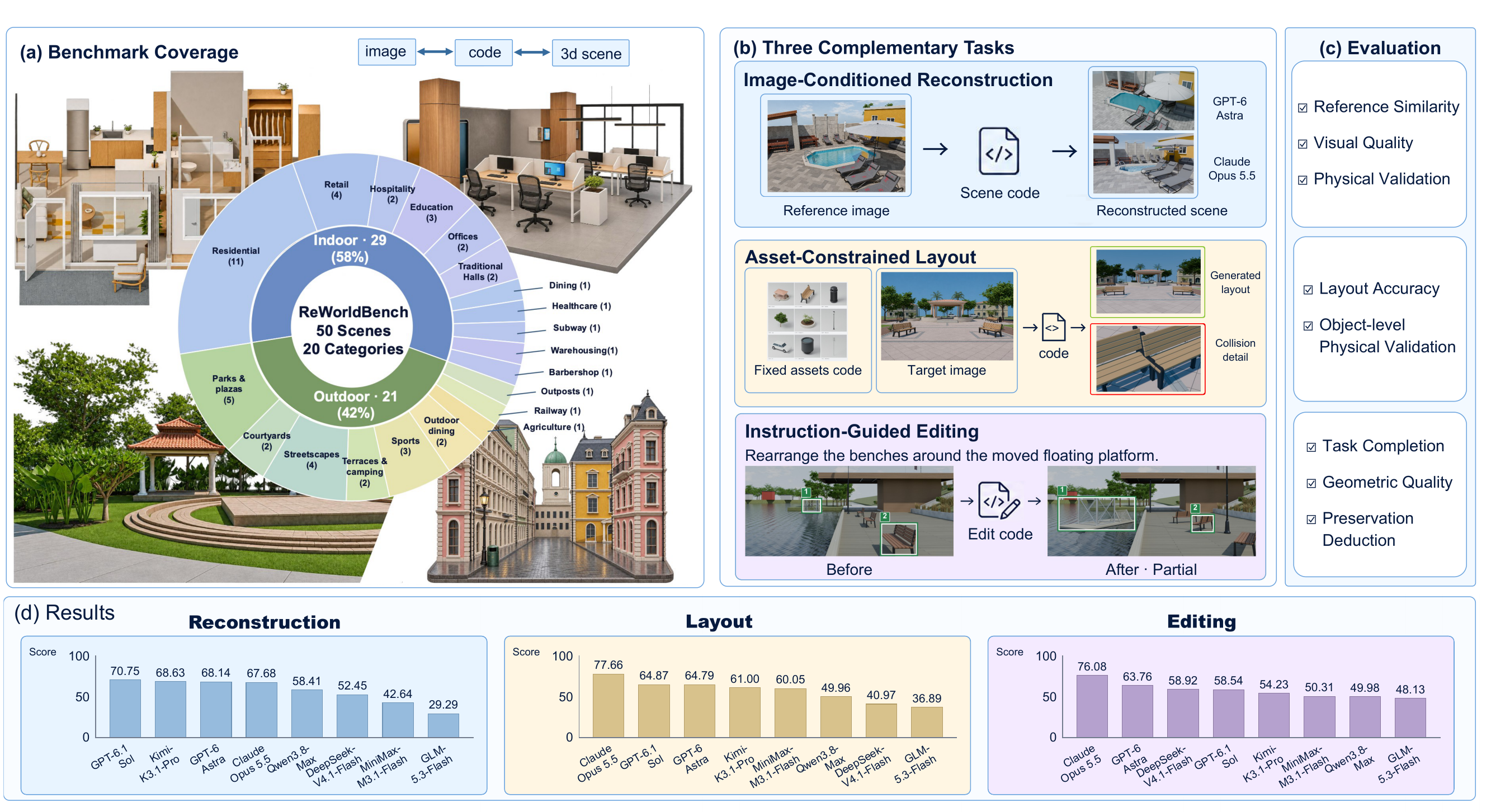}
\captionsetup{hypcap=false}
\captionof{figure}{\textbf{Overview of ReWorldBench.}
(a) Coverage of 50 indoor and outdoor scenes across 20 categories.
(b) Three complementary tasks: image-conditioned reconstruction,
asset-constrained layout, and instruction-guided editing.
(c) Task-specific evaluation of visual, spatial, and physical
quality, together with task fulfillment.
(d) Overall scores across eight coding agents on reconstruction, layout, and editing.}
\captionsetup{hypcap=true}
\label{fig:teaser}
\end{strip}
\begin{abstract}
Reliable 3D scenes provide a foundation for interactive experiences and embodied simulation. Coding agents offer a flexible way
to build and edit these scenes through executable programs
that describe geometry, appearance, and layout.
These tasks, however, require not only executable code and
convincing renderings, but also coherent layouts and physically
plausible object relationships. Assessing these requirements
is challenging, since spatial relationships are complex and
meshes are often not clearly grouped into objects. We introduce ReWorldBench, a benchmark for code-based
reconstruction and editing of realistic 3D scenes.
First, we construct a dataset using a multi-agent pipeline
for reference-image analysis and a three-stage workflow for
Blender Python code generation, yielding 50 scenes across
20 categories with 2,350 assets.
Second, we define three complementary tasks:
image-conditioned reconstruction, asset-constrained layout,
and instruction-guided editing. Together, they comprise
200 task instances spanning indoor and outdoor environments.
Third, we establish a task-specific evaluation framework
that assesses execution validity, task fulfillment,
visual quality, and physical plausibility.
Experiments on eight frontier coding agents show that, while leading agents reliably produce executable code, no single agent excels across all three tasks. These results highlight the gap between executable code and visually faithful, physically plausible, and controllable 3D scenes.

\end{abstract}    

\section{Introduction}
\label{sec:intro}

Realistic 3D scenes are increasingly in demand for interior
design, virtual environments, and embodied simulation~\cite{Holodeck,ProcTHOR,InfinigenIndoors}.
Constructing and editing these scenes requires translating visual references and user instructions~\cite{UNIM} into geometry, appearance, and spatial relationships~\cite{MMSIBench,MindCube}. With advances in software engineering and command-line tasks~\cite{jimenez2024swe,merrill2026terminal},
a promising approach is to use coding agents to construct
and modify 3D scenes through executable programs~\cite{ThreeDGPT,LL3M, li2026videococo}. Unlike black-box 3D generation methods with limited access
to scene structure, code-based construction explicitly
represents geometry, appearance, and layout, allowing agents
to inspect and modify individual objects and their spatial
relationships~\cite{MeshCoder,SceneScript,Procedura}.

However, successful execution and convincing renderings do not
necessarily indicate correct object placement or physically
plausible relationships.
Existing benchmarks progress from object modeling~\cite{CADPrompt,3DCodeBench} and program editing~\cite{BlenderGym,3DCodeEdit} to scene-level tasks.
WorldCoderBench~\cite{WorldCoder-Bench} and VoxelCodeBench~\cite{VoxelCodeBench} evaluate scene generation in simplified or voxelized settings with limited complexity and realism.
LEGO-Bench~\cite{LEGO-Bench} targets realistic scenes yet does not evaluate physical interactions, while BlenderWorld~\cite{BlenderWorld} introduces task-specific rubrics for spatial and visual relationships but lacks ground-truth scene code.
Code4Scene~\cite{Code4Scene} incorporates programmatic physical checks, though generic collision and static support tests remain insufficient to capture semantic physical plausibility.
Complementary benchmarks focus on physical executability in dynamic simulations and interactive mechanisms~\cite{PhysCodeBench,4DBuildBench,4DCodeBench,ExeWorldBench}.

\begin{figure}[!htbp]
    \centering
    \includegraphics[width=\linewidth]{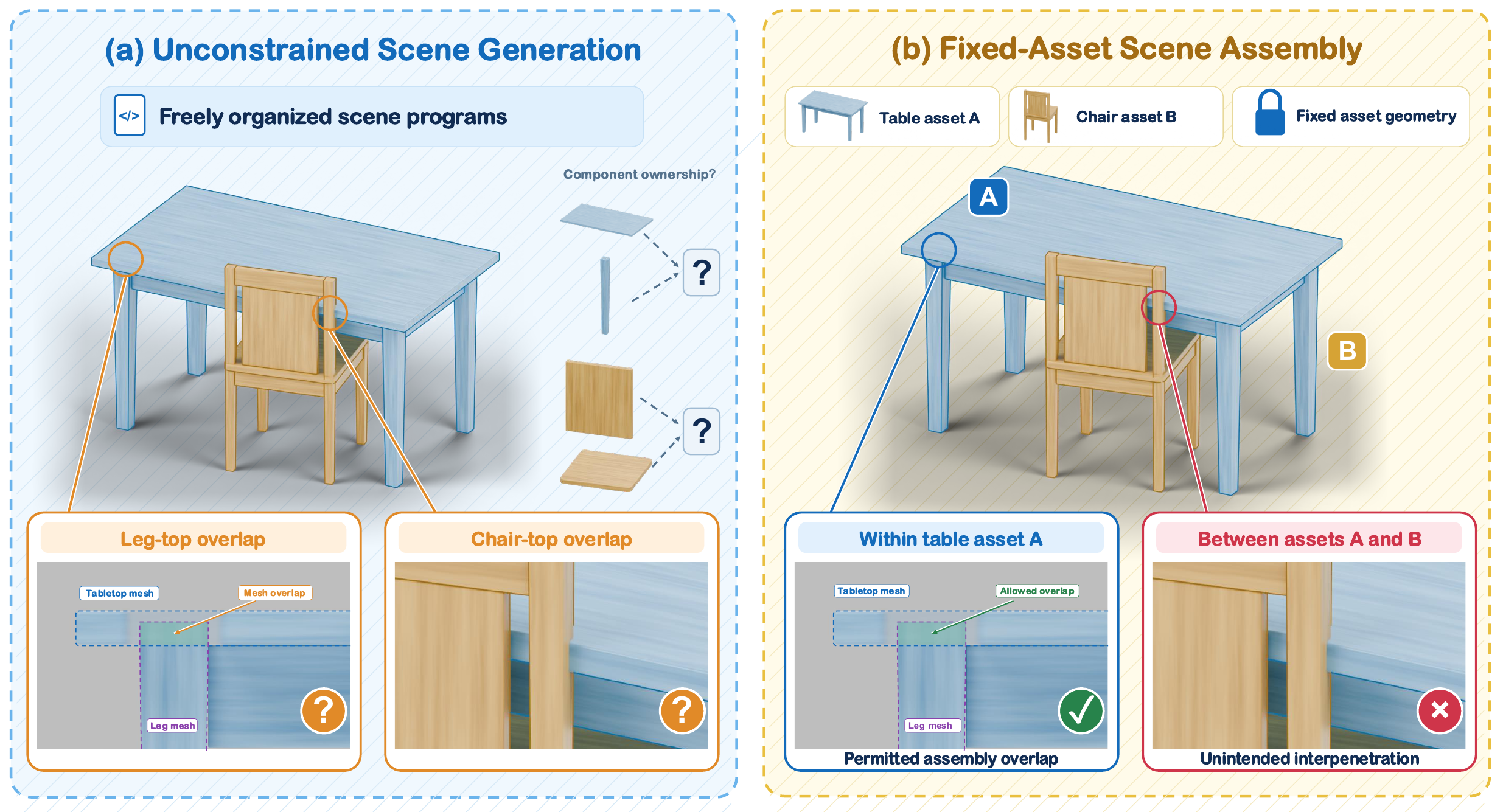}
    \caption{
        Why explicit asset boundaries matter.
        (a) Unknown component ownership leaves part assembly and object
        collision indistinguishable.
        (b) Fixed assets separate permitted internal overlap from
        unintended interpenetration.
    }
    \label{fig:physical_validation}
    \vspace{-5pt}
\end{figure}

For realistic scene reconstruction, physical validation is particularly challenging, since freely
generated programs vary in code organization and geometric
decomposition, often leaving object membership and intended
inter-object relationships implicit.
In such unconstrained programs (Fig.~\ref{fig:physical_validation}a),
unknown component ownership makes mesh overlap ambiguous: a table leg
overlapping its tabletop for valid assembly and a chair backrest
penetrating the tabletop both appear as intersecting meshes.
When scenes are instead arranged using fixed-geometry assets with
explicit boundaries (Fig.~\ref{fig:physical_validation}b), permitted
within-asset overlap can be separated from unintended inter-asset
penetration, enabling reliable object-level checks under physical
rules.

To address this gap, we introduce \textbf{ReWorldBench},
a benchmark for code-based reconstruction and editing of
realistic 3D scenes.
Our data construction pipeline produces 50 scenes across
20 categories, including 29 indoor and 21 outdoor environments,
with 2,350 assets.
It pairs reference images with executable Blender Python
programs and 3D scenes.
We independently model each semantic object as a reusable
asset composed of one or more meshes, with explicit component membership.

\begin{table*}[t]
\centering
\caption{\textbf{Comparison of representative 3D benchmarks.} $\checkmark$: reported; ---: not reported. ReWorldBench covers reconstruction, layout, and editing across indoor and outdoor scenes, combining task-specific rubrics with object-level physical checks and agent-based validation.}
\label{tab:benchmark_comparison}
\begingroup
\setlength{\tabcolsep}{4pt}
\renewcommand{\arraystretch}{1.15}
\noindent\resizebox{\linewidth}{!}{%
\begin{tabular}{@{}lccccccc@{}}
\toprule
& \multicolumn{3}{c}{\textbf{Setting}} & \multicolumn{1}{c}{\textbf{Data}} & \multicolumn{3}{c}{\textbf{Evaluation}} \\
\cmidrule(lr){2-4}\cmidrule(lr){5-5}\cmidrule(l){6-8}
\textbf{Benchmark} & \textbf{Platform} & \textbf{Task} & \shortstack{\textbf{Indoor/Outdoor}\\\textbf{Coverage}} & \shortstack{\textbf{GT}\\\textbf{Level}} & \textbf{Visual} & \shortstack{\textbf{Physical}\\\textbf{validation}} & \textbf{Task-specific} \\
\midrule
3DCodeBench~\cite{3DCodeBench} & Blender & Recon. & \textemdash & Object & \ensuremath{\checkmark} & \textemdash & \textemdash \\
VoxelCodeBench~\cite{VoxelCodeBench} & Unreal & Recon. & \textemdash & \textemdash & \ensuremath{\checkmark} & \textemdash & \textemdash \\
WorldCoder-Bench~\cite{WorldCoder-Bench} & Three.js & Recon. & \textemdash & \textemdash & \textemdash & \ensuremath{\checkmark} & \ensuremath{\checkmark} \\
Code4Scene~\cite{Code4Scene} & Unreal & Recon. / Edit. & \ensuremath{\checkmark} & Scene & \ensuremath{\checkmark} & \ensuremath{\checkmark} & \ensuremath{\checkmark} \\
PhysCodeBench~\cite{PhysCodeBench} & Genesis / MuJoCo & Recon. & \textemdash & Sim. & \ensuremath{\checkmark} & \ensuremath{\checkmark} & \ensuremath{\checkmark} \\
CADPrompt~\cite{CADPrompt} & CadQuery & Recon. & \textemdash & Object & \textemdash & \textemdash & \textemdash \\
BlenderGym~\cite{BlenderGym} & Blender & Edit. & \textemdash & Scene & \ensuremath{\checkmark} & \textemdash & \textemdash \\
3DCodeEdit~\cite{3DCodeEdit} & Blender & Edit. & \textemdash & \textemdash & \ensuremath{\checkmark} & \ensuremath{\checkmark} & \ensuremath{\checkmark} \\
\midrule
\rowcolor[gray]{0.93}
\textbf{ReWorldBench (ours)} & Blender & \textbf{Recon. / Layout / Edit.} & \ensuremath{\checkmark} & \textbf{Asset + Scene} & \ensuremath{\checkmark} & Object + Agent & \ensuremath{\checkmark} \\
\bottomrule
\end{tabular}}
\endgroup

\vspace{-8pt}
\end{table*}

Built on this scene collection, ReWorldBench comprises
200 instances across three complementary tasks.
\textbf{Image-Conditioned Reconstruction} requires
generating complete scenes from visual references.
\textbf{Asset-Constrained Layout} assesses spatial arrangement
using supplied assets with fixed geometry.
\textbf{Instruction-Guided Editing} requires targeted code
changes while preserving unaffected content.
Evaluation combines execution checks, task-specific rubrics,
and visual judgments~\cite{VQAScore,VisionQ}. Explicit asset
membership supports rule-based physical checks for layout and
editing, while freely generated reconstructions are assessed
through rubric-guided agent judgments and auxiliary geometric
checks. Among eight frontier coding agents, GPT-6.1 Sol performs
best on reconstruction and Claude Opus 5.5 on layout and editing,
yet most still struggle with fine-grained visual fidelity and
physical consistency.

Our contributions are threefold:
\begin{itemize}
    \item We develop a \textbf{data construction pipeline} that produces 50 diverse 3D scenes, executable programs, and a reusable library of 2,350 independently modeled assets.
    
    \item We introduce \textbf{ReWorldBench}, covering
reconstruction, layout, and editing, with task-specific
evaluation and \textbf{object-level physical validation}
based on explicit object membership and physical rules.

    \item We conduct an \textbf{empirical evaluation of representative coding agents}, analyzing their strengths and failure modes across the three tasks.
\end{itemize}

\section{Related Work}
\label{sec:related_work}

\noindent\textbf{Benchmarks for 3D Programming.}\quad
Prior benchmarks cover CAD and object
generation~\cite{CADPrompt,3DCodeBench} and Blender
script editing~\cite{BlenderGym,3DCodeEdit}.
At a larger scale, recent works build or manipulate multi-object environments~\cite{VoxelCodeBench,WorldCoder-Bench,LEGO-Bench,SceneActBench,Code4Scene,BlenderWorld} and simulate dynamic mechanisms~\cite{PhysCodeBench,4DBuildBench,4DCodeBench,ExeWorldBench}. As shown in \cref{tab:benchmark_comparison}, ReWorldBench combines reconstruction, layout, and editing
across indoor and outdoor scenes, providing both asset and scene code along with visual, task-specific, and component-aware physical evaluation.

\noindent\textbf{3D Generation and Spatial Reasoning.}\quad
Structured representations support controllable 3D generation,
from part assemblies~\cite{ShapeAssembly,Procedura} and object
layouts~\cite{ATISS,LayoutGPT} to procedural
environments~\cite{Infinigen,InfinigenIndoors,Holodeck,ProcTHOR}.
Work on spatial reasoning examines how models understand
object relationships in visual
observations~\cite{VSR,SpatialVLM,SpatialRGPT,SAT}.
For code-based generation, scene graphs and staged workflows
guide the construction of executable
scenes~\cite{SceneCraft,Code-as-Room,ThreeDGPT}, while rendering and
execution feedback support iterative program
refinement~\cite{VIGA,LEGO-Bench,4DBuildBench,3DCodeVerse,BlenderAlchemy,LL3M}.
Our data pipeline combines structured asset modeling and scene
assembly with visual and physical verification.
We implement assets independently and retain explicit
component membership, yielding reusable asset programs
for scene construction and object-level evaluation.

\begin{figure*}[t]
    \centering
    \includegraphics[width=\textwidth]{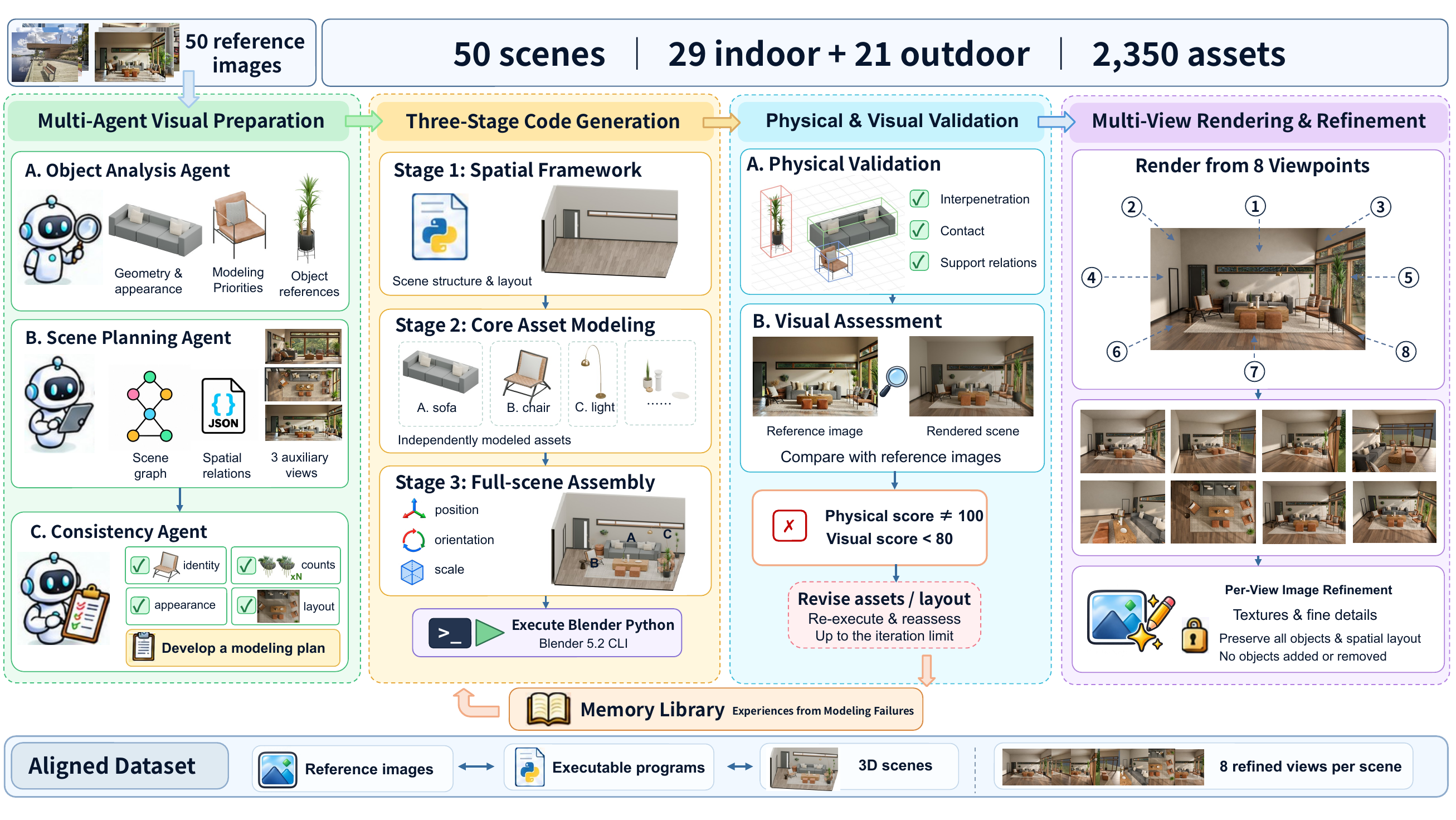}
    \caption{
         \textbf{Overview of the ReWorldBench data construction pipeline.}
        Reference images are transformed into executable 3D scenes
with reusable assets, followed by physical and
visual validation and multi-view image refinement.
    }
    \label{fig:data_pipeline}
    \vspace{-8pt}
\end{figure*}

\section{Data Construction}
\label{sec:dataset}
To construct a dataset of aligned code, scenes,
and images with explicit object boundaries and physical
relationships, we develop a four-stage pipeline
(\cref{fig:data_pipeline}).
The pipeline generates Blender Python programs from images and executes them through the Blender 5.2 CLI
to construct 3D scenes.

\subsection{Multi-Agent Visual Preparation}
\label{sec:visual_preparation}
Generating scene programs from reference images
requires understanding both object details and spatial
relationships. We use a multi-agent workflow to analyze
and consolidate this information into a consistent modeling plan.

\noindent\textbf{Object Analysis Agent.}\quad
The agent identifies key objects and describes their geometry
and appearance.
It assigns objects to primary, secondary, and tertiary tiers
according to their importance to reconstruction. For each object, it invokes an image generation model to
produce object-centric reference images for asset modeling.

\noindent\textbf{Scene Planning Agent.}\quad
Given the reference image and object inventory, the agent constructs a scene graph specifying architectural
structures, object instances, and spatial relationships.
It also invokes
an image generation model to synthesize three auxiliary views.
These offer plausible interpretations of scene geometry;
the original image remains the primary visual evidence.

\noindent\textbf{Consistency Agent.}\quad
The agent verifies the consistency of object descriptions, generated images, and the scene graph against the original reference, focusing on object identity, instance counts, appearance, and spatial arrangement. It then consolidates the verified information into a modeling
plan for code generation.

\subsection{Three-Stage Code Generation}
\label{sec:code_generation}
Based on the modeling plan, we progressively construct executable programs through three stages.

\noindent\textbf{Spatial Framework Construction.}\quad
The first stage establishes the architectural structure
and coarse spatial layout from the modeling plan
and visual references.

\noindent\textbf{Core Asset Modeling.}\quad
The second stage generates individual assets from object descriptions and object-centric reference images. Each asset is implemented independently for inspection
and refinement before scene assembly. The assets form
a reusable code library with explicit component membership.

\noindent\textbf{Full-Scene Assembly.}\quad
The third stage assembles the assets within the spatial
framework, setting their positions, orientations, and scales
according to the modeling plan. The generated program is then executed to construct the complete 3D scene.

\noindent\textbf{Memory Library.}\quad
Across these stages, we record recurring modeling failures
and corresponding solutions in a memory library.
These include modeling challenges such as detailed tree geometry,
foliage, and upholstered objects.
The generation agent consults relevant records when constructing
subsequent assets and scenes
(see Appendix~\ref{sec:supp_memory_mechanism}).

\subsection{Physical and Visual Validation}
\label{sec:construction_feedback}

We apply physical validation checks described in \cref{sec:evaluation},
together with visual assessments of the reconstructed scenes
against their reference images. The feedback guides the agent
in refining assets and correcting spatial arrangements.
After each revision, we execute the program and reassess
the scene until it achieves a full physical validation
score and a visual score above 80, or the revision limit
is reached. Only scenes meeting both criteria are retained.

\subsection{Multi-View Rendering and Image Refinement}
\label{sec:image_refinement}

For each retained scene, a coding agent renders eight views
in Blender, and an image generation model refines only the
surface textures.
We manually check fidelity to the original renders and cross-view consistency, retaining only images that preserve object identities, counts, geometry, and spatial relationships as task references.


\section{ReWorldBench}
\label{sec:benchmark}

\subsection{Benchmark Overview}
\label{sec:benchmark_overview}
\begin{figure*}[t]
    \centering
    \includegraphics[width=\textwidth]{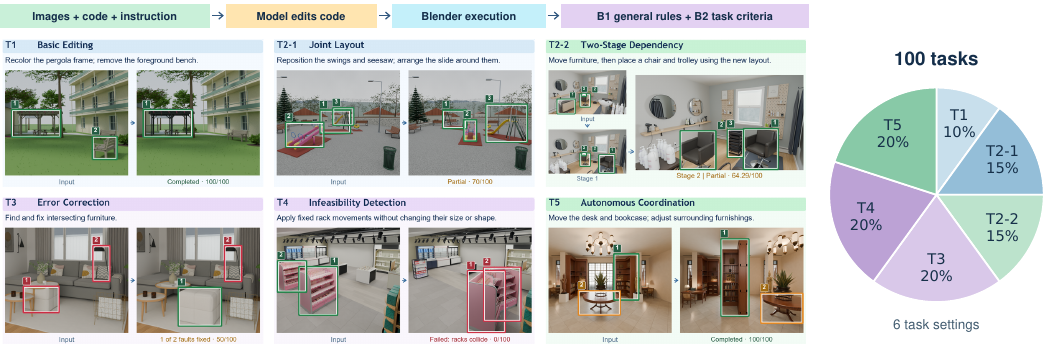}
    \caption{\textbf{Instruction-guided scene editing:} examples of six task settings and their distribution across 100 instances. T4 illustrates an invalid execution; the expected response is refusal.}
    \label{fig:editing_task_overview}
    \vspace{-10pt}
\end{figure*}

Building on the scene collection described in
\cref{sec:dataset}, ReWorldBench comprises 200 task
instances across three complementary tasks
(\cref{fig:teaser}(b)).
Image-conditioned reconstruction assesses the ability to
recover scene geometry, appearance, and layout from visual
references. Asset-constrained layout examines spatial
arrangement and physical relationships among supplied assets.
Instruction-guided editing evaluates targeted modifications
while preserving unaffected scene content.
For each task, agents generate or modify scene programs
that are executed in Blender. Evaluation combines execution checks with task-specific
metrics and rubrics to assess task fulfillment, visual quality,
layout accuracy, and physical plausibility
(\cref{fig:teaser}(c)).

\subsection{Task Design}
\label{sec:task_design}

We denote the coding agent by $\mathcal{A}$ and its generated
program by $\hat{P}$. Each task specifies its inputs and
constraints. Submitted programs are executed in Blender
and evaluated as described in \cref{sec:evaluation}.

\noindent\textbf{Image-Conditioned Reconstruction.}\quad
Given one or more reference views $\mathcal{I}$
of the same scene, the agent
generates a complete scene program:
\begin{equation}
    \hat{P}_{\mathrm{rec}} = \mathcal{A}(\mathcal{I}).
    \label{eq:reconstruction}
\end{equation}
The agent infers object geometry, appearance, and spatial
arrangement from the images and expresses them in code. Single-view reconstruction requires reasoning from
incomplete observations, while multi-view reconstruction
requires integrating evidence across viewpoints.

\noindent\textbf{Asset-Constrained Layout.}\quad
Given a target layout image $I$, a fixed asset
collection $\mathcal{B}$, and a basic architectural
structure $H$, the agent generates a scene program:
\begin{equation}
    \hat{P}_{\mathrm{lay}} =
    \mathcal{A}(I, \mathcal{B}, H).
    \label{eq:layout}
\end{equation}
The agent adjusts asset positions and orientations within
$H$ to match the target image while preserving asset geometry
and the architectural structure. At the same time, it must
maintain valid support and attachment relationships without
unsupported floating or unintended interpenetration.

\begin{table*}[t]
\centering
\caption{Performance across the three ReWorldBench tasks
    (0--100; higher is better).
    Bold indicates the best result, and underlining indicates
    the second-best result in each column.}
\label{tab:main_results}
\fontsize{7.5}{9}\selectfont
\setlength{\tabcolsep}{3.2pt}\renewcommand{\arraystretch}{1.22}
\begin{tabular}{@{}lrrrrrrrrrrrrrrrr@{}}
\toprule
& \multicolumn{5}{c}{Image-Conditioned Reconstruction} & \multicolumn{4}{c}{Asset-Constrained Layout} & \multicolumn{7}{c}{Instruction-Guided Editing} \\
\cmidrule(lr){2-6}\cmidrule(lr){7-10}\cmidrule(l){11-17}
Model & E$\uparrow$ & S$\uparrow$ & V$\uparrow$ & P$\uparrow$ & \cellcolor{scoreblue}\textbf{Score}$\uparrow$ & E$\uparrow$ & A$\uparrow$ & P$\uparrow$ & \cellcolor{scoreyellow}\textbf{Score}$\uparrow$ & T1$\uparrow$ & T2-1$\uparrow$ & T2-2$\uparrow$ & T3$\uparrow$ & T4$\uparrow$ & T5$\uparrow$ & \cellcolor{scoregreen}\textbf{Score}$\uparrow$ \\
\midrule
\modelicon{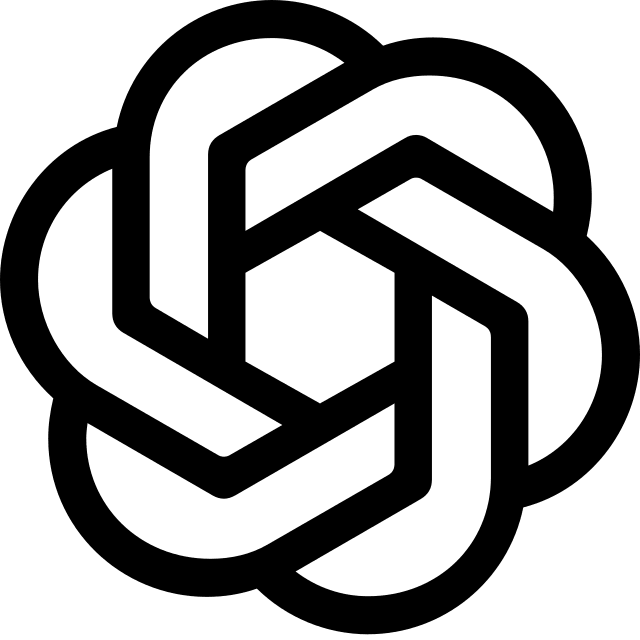}GPT-6 Astra & 96.00 & \underline{62.92} & 54.53 & 88.71 & \cellcolor{scoreblue}68.14 & \textbf{98.00} & 70.43 & \underline{59.15} & \cellcolor{scoreyellow}64.79 & 72.50 & \underline{70.77} & \textbf{80.09} & \underline{40.00} & 45.00 & \underline{84.39} & \cellcolor{scoregreen}\underline{63.76} \\
\modelicon{openai}GPT-6.1 Sol & \textbf{100.00} & \textbf{65.21} & \textbf{56.33} & \textbf{92.57} & \cellcolor{scoreblue}\textbf{70.75} & \textbf{98.00} & \underline{71.23} & 58.51 & \cellcolor{scoreyellow}\underline{64.87} & \textbf{90.00} & \textbf{71.42} & 55.31 & 37.00 & 35.00 & 80.64 & \cellcolor{scoregreen}58.54 \\
\modelicon{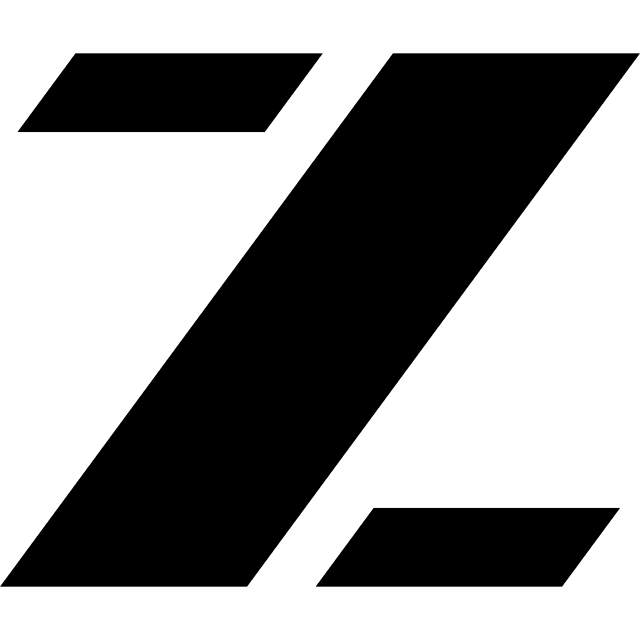}GLM-5.3-Flash & 52.00 & 26.93 & 23.49 & 38.24 & \cellcolor{scoreblue}29.29 & 78.00 & 41.90 & 31.89 & \cellcolor{scoreyellow}36.89 & 74.00 & 56.34 & 69.93 & 36.50 & 5.00 & 67.45 & \cellcolor{scoregreen}48.13 \\
\modelicon{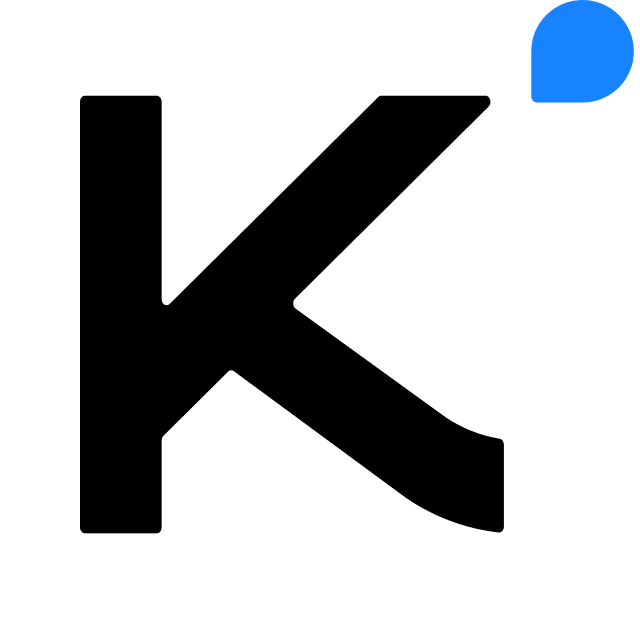}Kimi-K3.1-Pro & \underline{98.00} & 61.79 & \underline{55.30} & 91.08 & \cellcolor{scoreblue}\underline{68.63} & \underline{96.00} & 71.05 & 50.95 & \cellcolor{scoreyellow}61.00 & 81.50 & 66.63 & \underline{73.84} & 32.00 & 15.00 & 78.04 & \cellcolor{scoregreen}54.23 \\
\modelicon{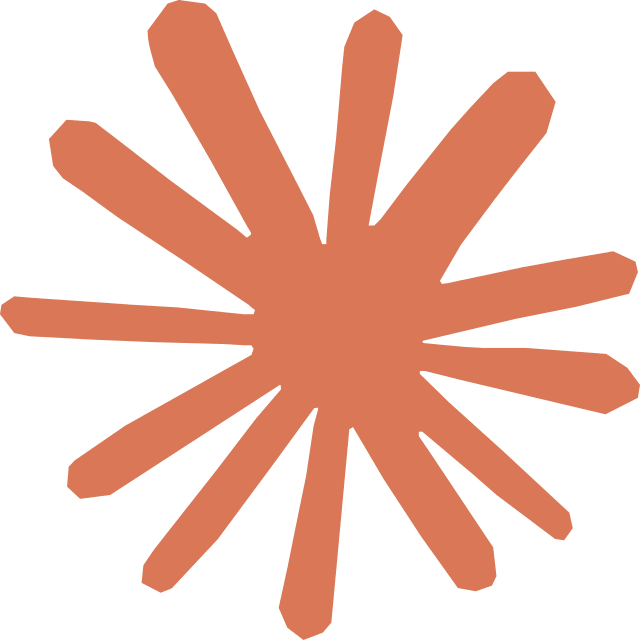}Claude Opus 5.5 & 94.00 & 61.54 & 51.40 & \underline{92.15} & \cellcolor{scoreblue}67.68 & \textbf{98.00} & \textbf{87.68} & \textbf{67.63} & \cellcolor{scoreyellow}\textbf{77.66} & 84.00 & 62.18 & 66.38 & \textbf{61.50} & \textbf{95.00} & \textbf{85.50} & \cellcolor{scoregreen}\textbf{76.08} \\
\modelicon{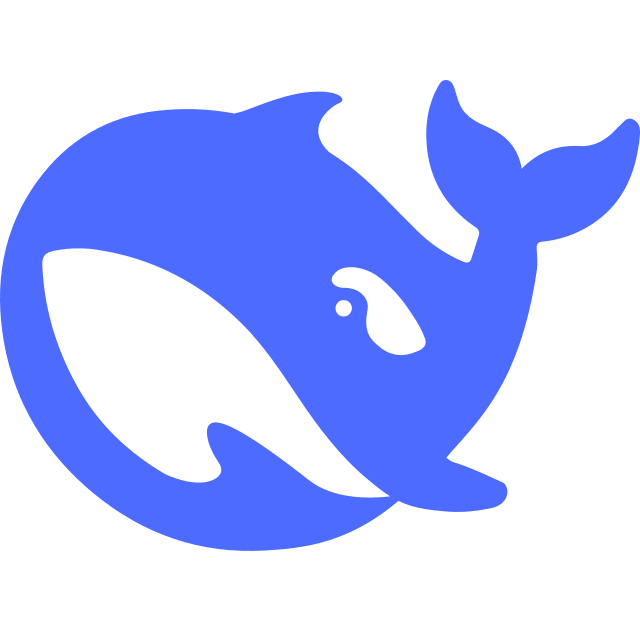}DeepSeek-V4.1-Flash & 80.00 & 46.02 & 42.18 & 71.28 & \cellcolor{scoreblue}52.45 & 72.00 & 49.14 & 32.81 & \cellcolor{scoreyellow}40.97 & \underline{88.00} & 54.90 & 63.07 & 29.50 & \underline{50.00} & 82.64 & \cellcolor{scoregreen}58.92 \\
\modelicon{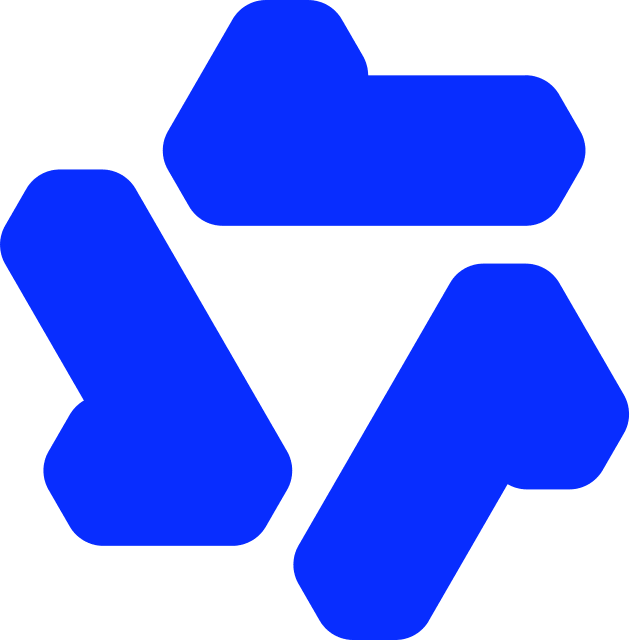}Qwen3.8-Max & 88.00 & 53.68 & 43.62 & 79.51 & \cellcolor{scoreblue}58.41 & \underline{96.00} & 57.20 & 42.72 & \cellcolor{scoreyellow}49.96 & 86.50 & 60.63 & 70.80 & 32.00 & 5.00 & 71.07 & \cellcolor{scoregreen}49.98 \\
\modelicon{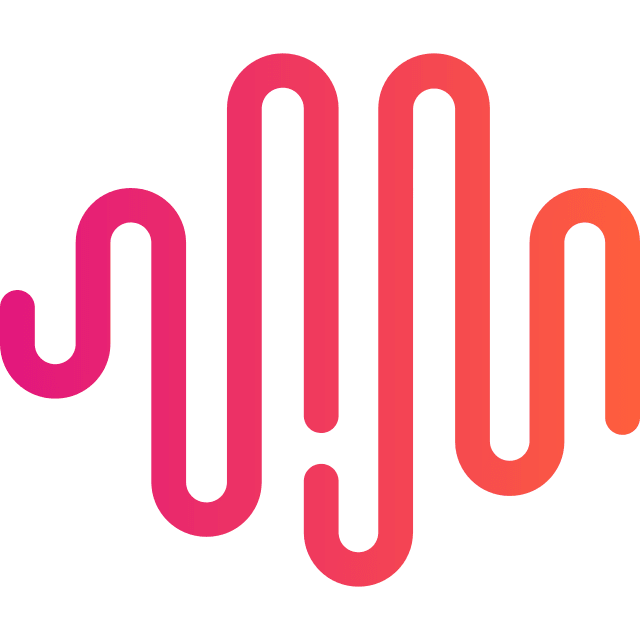}MiniMax-M3.1-Flash & 64.00 & 38.37 & 33.24 & 57.72 & \cellcolor{scoreblue}42.64 & \underline{96.00} & 67.81 & 52.29 & \cellcolor{scoreyellow}60.05 & 75.00 & 62.61 & 59.63 & 29.00 & 20.00 & 73.35 & \cellcolor{scoregreen}50.31 \\
\bottomrule\end{tabular}
\vspace{-8pt}
\end{table*}

\noindent\textbf{Instruction-Guided Editing.}\quad
Given an initial scene program $P_0$ and instruction $u$, the agent returns
\begin{equation}
\hat P_{\mathrm{edit}}=\mathcal A(P_0,u),
\label{eq:editing}
\end{equation}
where the output is an executable edited program or a refusal
if the request is infeasible. The task assesses instruction following, coordination of
affected objects, and preservation of unrelated content.

We design six editing settings around three capabilities
(\cref{fig:editing_task_overview}).
\textit{Instruction execution} includes basic compositional
editing (T1), which applies different operations to distinct
assets, and joint multi-asset editing (T2-1), which combines
numerical transformations with relative spatial constraints.
\textit{Dependency maintenance} includes multi-stage dependent
editing (T2-2), where both stages are provided upfront and
later operations use the actual preceding scene state,
and autonomous conflict resolution (T5), where agents identify
and rearrange obstructing assets without a complete obstacle list.
\textit{Scene diagnosis and feasibility judgment} includes
geometric fault repair (T3), requiring agents to locate and
correct injected defects, and infeasibility judgment (T4),
requiring them to identify conflicting constraints and
refuse infeasible requests.
Except for T3, all editing prompts explicitly allow refusal when
an operation is infeasible, giving all models the same refusal option.
Only correct refusals in T4 earn credit; all other refusals score zero.

\subsection{Evaluation and Metrics}
\label{sec:evaluation}

\noindent\textbf{Code Execution and Output Validity.}\quad
We check whether code generation finishes within the time
limit and whether the program executes without errors
and produces the required outputs.
Each program submission receives a validity indicator
$E_i \in \{0,1\}$. For reconstruction and layout, invalid outputs score zero;
editing follows the task-specific scoring rules below.
The infeasibility judgment setting (T4) in editing instead
evaluates refusal correctness without requiring code.

\noindent\textbf{Image-Conditioned Reconstruction.}\quad
We use an automated agent to configure cameras
and render each reconstructed scene from four viewpoints.
We then assess the scene using a task-specific rubric covering
reference similarity ($S$), visual quality ($V$), and
physical validation ($P$).

We use a VLM judge for $S$ and $V$.
Reference similarity compares candidate renders with ground-truth
images in object content, layout, geometry, and material appearance.
Visual quality assesses modeling craftsmanship, materials and
textures, and visual coherence without reference images.

For physical validation, we use an agent judge.
Guided by the rubric, it interprets object identities and
relationships in the images to identify potential physical
violations. It then investigates suspected issues using
auxiliary Blender checks and additional views when needed.

We normalize each dimension to $[0,100]$ using rubric item
weights and compute
\begin{equation}
    R_i = E_i \left(0.40S_i + 0.30V_i + 0.30P_i\right).
    \label{eq:reconstruction_score}
\end{equation}
Each judgment includes supporting evidence and a concise rationale.

\noindent\textbf{Asset-Constrained Layout.}\quad
We assess layout accuracy and perform object-level physical
validation. Unlike freely generated reconstructions, the
supplied assets have explicit component membership,
enabling direct rule-based checks.
We use collision and contact checks in Blender to detect
unintended interpenetration, unsupported floating, and
incorrect attachment under the task-specific rubric.

We compare the layout with the ground-truth scene
and score asset placement, orientation, and instance counts
according to the rubric. We normalize physical validation
($P_i$) and layout accuracy ($A_i$) to $[0,100]$ and compute
\begin{equation}
    L_i = E_i \left(0.50P_i + 0.50A_i\right).
    \label{eq:layout_score}
\end{equation}

\noindent\textbf{Instruction-Guided Editing.}\quad
We evaluate three dimensions:
(1) task completion measures fulfillment of requested operations
and spatial relations;
(2) physical validation applies the checks from asset-constrained
layout to detect interpenetration and support violations,
comparing pre- and post-edit scenes to distinguish
pre-existing defects from those introduced by editing; and
(3) scene preservation assesses asset integrity and unaffected
content. For T1, T2-1, and T5, we compute
\begin{equation}
    Q_i=\max\!\left(0,50C_i+50\min(C_i,P_i)-D_i\right),
    \label{eq:editing_score}
\end{equation}
where $C_i,P_i\in[0,1]$ denote task completion and physical
validation scores, respectively, and $D_i$ is a point deduction
for preservation violations. Physical validation credit is
capped by completion to prevent unwarranted credit for
unchanged scenes.
For T2-2, we score both checkpoints and
aggregate their results. T3 assesses fault repair and
preservation; T4 assesses correct refusal of infeasible requests. Detailed scoring rules are provided in Appendix~\ref{sec:editing_scoring_details}.

\noindent\textbf{Rubric Construction.}\quad
For reference-similarity and visual-quality assessment,
we derive criteria from scene characteristics, reference
images, asset inventories, and scene graphs.
We record each criterion's source and define scoring anchors
from 0 (absent or unrecognizable) to 6 (fully satisfied); for physical validation, we construct scene-specific rubrics from asset inventories and scene graphs to define support and attachment relations, such as a lamp resting on a table or a frame mounted on a wall, with explicit contact and overlap tolerances; for layout accuracy, we extract target asset positions and
orientations from the ground-truth scene in a fixed
architectural coordinate frame. The rubric assigns equal
weights to position and orientation scores and specifies
penalties for missing or extra instances. For editing, we use the instruction and initial scene to
define shared requirements (B1) for physical validity and
scene preservation, and task-specific criteria (B2) for
the requested operations and target states.

We pilot each rubric on outputs from three randomly selected
models and manually review the criteria and judgments.
We revise or remove unsupported, ambiguous, or unassessable
criteria and repeat the pilot evaluation.
We accept a rubric only when all three models produce complete,
evidence-supported judgments and review issues are resolved.
Details and examples appear in Appendix~\ref{sec:supp_judges}.

\begin{figure*}[t]
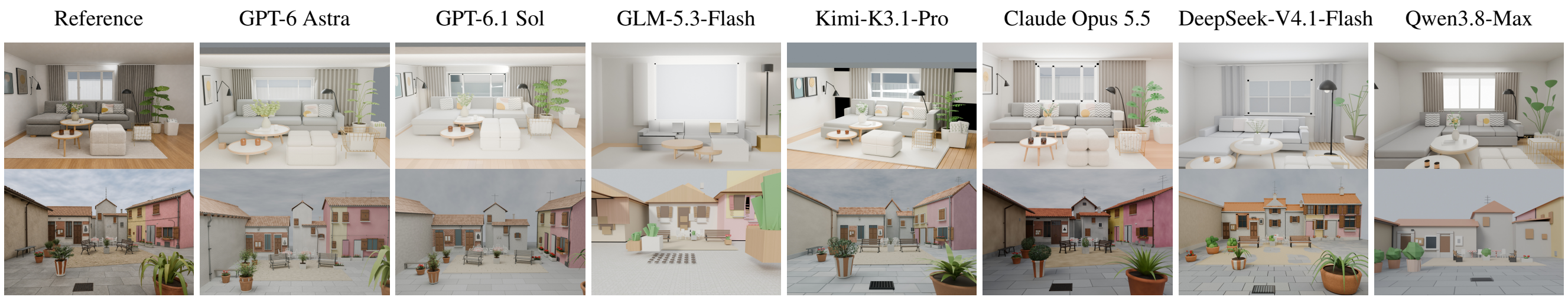

    \centering
    \setlength{\tabcolsep}{1pt}

    \edef\comparisonwidth{\the\dimexpr(\textwidth-14\tabcolsep)/8\relax}
    \edef\comparisonheight{\the\dimexpr(\textwidth-14\tabcolsep)/12\relax}

    \newcommand{\comparisoncell}[3]{%
        \begin{minipage}[c][\comparisonheight][c]{\comparisonwidth}
            \centering
            \includegraphics[width=\linewidth,height=\comparisonheight]{#2/#3}%
        \end{minipage}%
    }

    \newcommand{\modelheading}[1]{%
        \begin{minipage}[b]{\comparisonwidth}
            \centering
            \fontsize{7}{8}\selectfont
            \mbox{#1}\strut
        \end{minipage}%
    }

    \newcommand{\indoorcell}[1]{%
        \comparisoncell{b}{fig/task1-indoor-026}{#1}%
    }
    \newcommand{\outdoorcell}[1]{%
        \comparisoncell{t}{fig/task1-outdoor-015}{#1}%
    }

    \begin{tabular}{@{}cccccccc@{}}
        \modelheading{Reference}
        & \modelheading{GPT-6 Astra}
        & \modelheading{GPT-6.1 Sol}
        & \modelheading{GLM-5.3-Flash}
        & \modelheading{Kimi-K3.1-Pro}
        & \modelheading{Claude Opus 5.5}
        & \modelheading{DeepSeek-V4.1-Flash}
        & \modelheading{Qwen3.8-Max} \\[2pt]

        \indoorcell{reference.jpg}
        & \indoorcell{astra.jpg}
        & \indoorcell{sol.jpg}
        & \indoorcell{glm.jpg}
        & \indoorcell{kimi.jpg}
        & \indoorcell{opus.jpg}
        & \indoorcell{deepseek.jpg}
        & \indoorcell{qwen.jpg} \\[1pt]

        \outdoorcell{reference.jpg}
        & \outdoorcell{astra.jpg}
        & \outdoorcell{sol.jpg}
        & \outdoorcell{glm.jpg}
        & \outdoorcell{kimi.jpg}
        & \outdoorcell{opus.jpg}
        & \outdoorcell{deepseek.jpg}
        & \outdoorcell{qwen.jpg}
    \end{tabular}

    \caption{
        Qualitative results for image-conditioned reconstruction on indoor (top) and outdoor (bottom) scenes, with references
in the first column. MiniMax-M3.1-Flash submitted no valid code
for either scene and is omitted.
    }
    \label{fig:task1_qualitative}
    \vspace{-8pt}
\end{figure*}
\section{Experiments}
\label{sec:experiments}
\subsection{Experimental Setup}
\label{sec:experimental_setup}

\noindent\textbf{Models and Environment.}\quad
We evaluate coding agents powered by eight models:
GPT-6 Astra~\cite{OpenAIAstra2026}, GPT-6.1 Sol~\cite{OpenAISol2026}, GLM-5.3-Flash~\cite{GLMFlash2026}, Kimi-K3.1-Pro~\cite{KimiPro2026},
Claude Opus 5.5~\cite{Opus2026}, DeepSeek-V4.1-Flash~\cite{DeepSeekFlash2026}, Qwen3.8-Max~\cite{QwenMax2026},
and MiniMax-M3.1-Flash~\cite{MiniMaxFlash2026}.
Each agent has a 3,600-second budget per task instance
to produce a Blender Python program or, for infeasible
editing requests, a refusal.
We execute submitted programs using the Blender 5.2 CLI
to construct and render scenes.
Following \cref{sec:evaluation}, we report overall and
per-dimension scores averaged over all instances of each task.
Agent configurations and additional experimental settings
are provided in Appendix~\ref{sec:supp_implementation}.


\subsection{Main Results}
\label{sec:main_results}
\noindent\textbf{Image-Conditioned Reconstruction.}\quad
As shown in Table~\ref{tab:main_results}, GPT-6.1 Sol leads
in overall performance and all evaluation dimensions.
Kimi-K3.1-Pro, GPT-6 Astra, and Claude Opus 5.5 form a
closely matched group behind it, followed by Qwen3.8-Max
and DeepSeek-V4.1-Flash.
MiniMax-M3.1-Flash and GLM-5.3-Flash exhibit greater
limitations in execution reliability and visual fidelity.
Despite high output validity among the leading models,
reference similarity and visual quality remain limited,
with the best scores reaching only 65.21 and 56.33,
respectively.

\noindent\textbf{Asset-Constrained Layout.}\quad
Claude Opus 5.5 achieves the highest score of 77.66, leading in both layout accuracy (87.68) and physical validation (67.63). GPT-6.1 Sol and GPT-6 Astra follow at 64.87 and 64.79, respectively, with Sol performing slightly better in layout accuracy and Astra in physical validation. Across all models, physical validation scores are lower than layout accuracy scores, highlighting the difficulty of satisfying physical constraints even when asset placements broadly match the reference layout.

\noindent\textbf{Instruction-Guided Scene Editing.}\quad
Claude Opus 5.5 achieves the highest overall score of 76.08,
followed by GPT-6 Astra at 63.76.
Model strengths vary: GPT-6.1 Sol leads on T1 and T2-1,
GPT-6 Astra on T2-2, and Claude Opus 5.5 on T3--T5.
Despite all models scoring at least 72.50 on basic editing
(T1), fault repair (T3) scores reach only 61.50.
In infeasibility judgment (T4), Claude Opus 5.5 scores 95.00,
while all others score at most 50.00.
These results highlight a gap between performing explicit
edits and reliably repairing scene defects or recognizing
infeasible requests.

\begin{table}[t]
\centering
\caption{Execution failure counts across the three tasks.}
\label{tab:execution_failures}
\footnotesize
\setlength{\tabcolsep}{0pt}
\begin{tabular*}{\linewidth}{@{\extracolsep{\fill}}lrrr@{}}
\toprule
Model & Code errors & Agent failures & Total\\
\midrule
GPT-6 Astra & 6 & 1 & 7\\
GPT-6.1 Sol & 4 & 1 & 5\\
GLM-5.3-Flash & 34 & 14 & 48\\
Kimi-K3.1-Pro & 6 & 3 & 9\\
Claude Opus 5.5 & 3 & 5 & 8\\
DeepSeek-V4.1-Flash & 9 & 19 & 28\\
Qwen3.8-Max & 11 & 2 & 13\\
MiniMax-M3.1-Flash & 30 & 1 & 31\\
\bottomrule
\end{tabular*}
\vspace{-10pt}
\end{table}

\begin{figure}[t]
    \centering
    \includegraphics[width=\columnwidth]
    {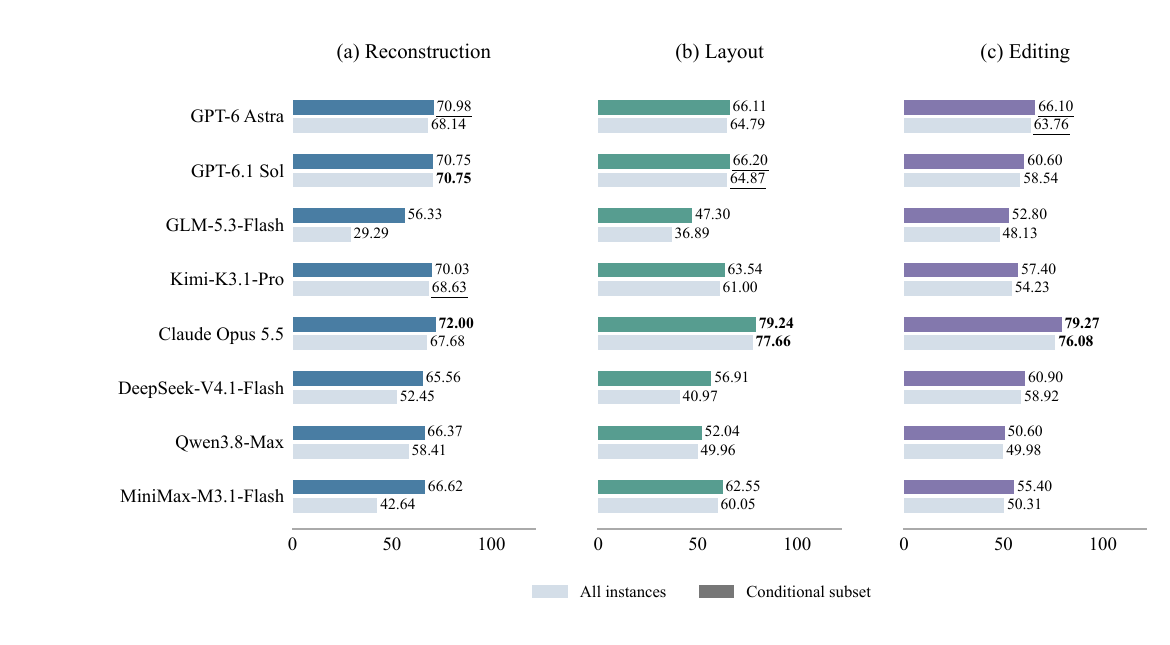}
    \caption{Overall and conditional scores (0--100). Conditional subsets retain successful reconstruction/layout outputs and editing responses without code or submission errors, including refusals. Conditional scores average retained instances.}
    \label{fig:error_analysis}
    \vspace{-10pt}
\end{figure}

\subsection{Failure and Limitation Analysis}
\label{sec:error_analysis}
\noindent\textbf{Execution Failures.}\quad
We distinguish code errors, including Python syntax errors
and incorrect Blender API usage, from agent failures,
including timeouts and invalid submission formats.
Their prevalence varies across models and tasks (\cref{tab:execution_failures}). Code correctness is the main execution bottleneck for
GLM-5.3-Flash and MiniMax-M3.1-Flash. DeepSeek-V4.1-Flash instead incurs 19 agent failures. Inspection of its trajectories
shows that prolonged analysis often exhausts the time
budget before submission.
The remaining models exhibit relatively high execution
reliability, with substantially fewer failures.
Figure~\ref{fig:error_analysis} compares scores across
all instances with those excluding execution failures.

\noindent\textbf{Visual Fidelity and Detail.}\quad
The qualitative comparisons in \cref{fig:task1_qualitative} show that recognizable composition does not ensure fine-grained visual fidelity. GPT-6 Astra, GPT-6.1 Sol, Claude Opus 5.5,
and Kimi-K3.1-Pro produce stronger visual results, but still
simplify upholstered furniture contours and surface folds. Outputs also differ in their reproduction of furniture
structures, lighting fixtures, and surface materials.
In the outdoor example, these four models preserve finer
branches and foliage, whereas others substitute coarse
geometric forms. These observations reveal a gap between recovering major scene elements and faithfully reproducing local shape and appearance.

\noindent\textbf{Spatial Configuration.}\quad
Recovering major objects does not guarantee accurate spatial relationships.
In reconstruction examples, models tend to align obliquely placed objects
with surrounding structures. Even with fixed assets, they may miss
orientation requirements despite approximately correct positions.
Similar difficulties arise in joint multi-asset editing (T2-1):
Qwen3.8-Max translated and rotated the sofa and armchair, but failed
to center the coffee table in front of the transformed sofa and align
it parallel to the sofa. This suggests that completing individual
object transformations does not guarantee correct spatial relationships
between objects.

\noindent\textbf{Physical Plausibility.}\quad
Models often reconstruct recognizable objects while overlooking
support, contact, and clearance requirements.
Visually plausible objects therefore do not ensure physically
consistent scenes. This limitation also appears in fault repair:
some models restore support for a floating chair but leave
another interpenetration unresolved. Such partial repairs
suggest difficulty in checking physical constraints across
the scene beyond individual objects.

\noindent\textbf{Fault Repair and Infeasibility Judgment.}\quad
Editing capabilities vary across models. All eight models score at least 72.50 on basic editing (T1),
but this strength does not consistently extend to fault
repair or infeasibility judgment. Claude Opus 5.5 leads
T3 and T4 with scores of 61.50 and 95.00, respectively.
Elsewhere, DeepSeek-V4.1-Flash is competitive on T5,
Kimi-K3.1-Pro outperforms Claude Opus 5.5 on T2-2,
and GPT-6.1 Sol and GPT-6 Astra lead T2-1 and T2-2,
respectively. These results suggest that basic editing
and multi-object coordination do not ensure reliable
fault repair or feasibility judgment.

\subsection{Evaluation across Settings}
\label{sec:task_configurations}
\begin{figure}[t]
\centering
\includegraphics[width=\columnwidth]{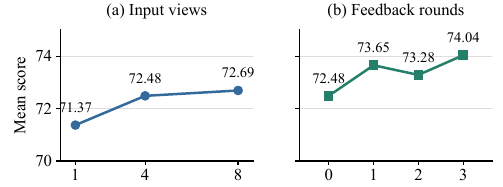}
\caption{Effects of input views and visual feedback on Task 1
using GPT-6.1 Sol on 20 scenes.
Round 0 is the four-view baseline without feedback.}
\label{fig:settings_ablation}
\vspace{-10pt}
\end{figure}
We evaluate GPT-6.1 Sol on 20 scenes to examine
how input views and visual feedback affect reconstruction
quality (\cref{fig:settings_ablation}).

\noindent\textbf{Single-View versus Multi-View Inputs.}\quad
Additional views provide modest average gains.
The mean score increases from 71.37 with one view to
72.48 with four views and 72.69 with eight views,
with little further improvement beyond four views.
However, each multi-view setting scores below the
single-view baseline on nine scenes.
This may reflect a trade-off: additional views supply
more visual evidence but also increase the difficulty
of integrating that evidence into a coherent 3D scene.

\noindent\textbf{Effect of Visual Feedback.}\quad
We let the agent inspect its rendered images after each
round, identify issues to improve, and revise its code in
the next round.
Visual feedback improves the mean score from 72.48 to 74.04
after three rounds, but the gains are not monotonic:
the score drops from 73.65 after the first round to 73.28
after the second. Seven scenes also finish below their
initial scores.
Inspection of agent trajectories
(see Appendix~\ref{sec:feedback_trajectories}) shows that
agents often identify visual discrepancies but fail to
correct them or make only minor adjustments.
They frequently adjust lighting while leaving the identified
geometry or layout errors unresolved.
This reveals a gap between recognizing visual errors and
making targeted code changes that fix the underlying scene
structure.

\subsection{Agreement with Human Evaluation}
\label{sec:human_evaluation}


\begin{figure}[t]
    \centering
    \includegraphics[width=\columnwidth]
    {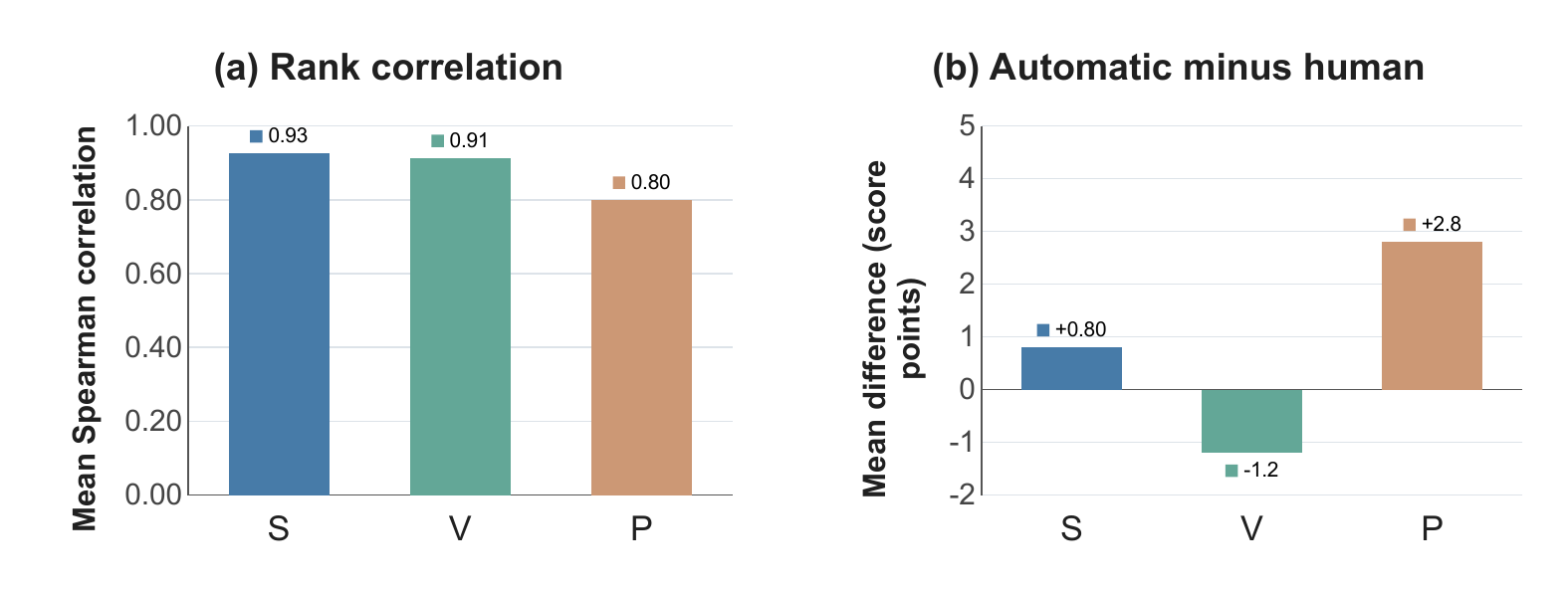}
    \caption{Agreement between automatic judges and human
    evaluators on Task~1 for $S$,
    $V$, and $P$.
    (a)~Spearman's correlation, averaged over three evaluators.
    (b)~Mean score difference (automatic minus human).}
    \label{fig:human_agreement}
    \vspace{-10pt}
\end{figure}

We randomly sample 10 Task~1 scenes and collect the outputs
of all eight agents. Each output is scored independently by
three human evaluators using the same rubrics as the
automatic judges. For $S$ and $V$, evaluators view the same
rendered images as the VLM judge; for $P$, they can also
inspect the scene in Blender. For each scene, we compute
Spearman's correlation between automatic and human scores
across the eight agents, and average it over scenes and
evaluators. As shown in Figure~\ref{fig:human_agreement},
the automatic scores agree well with human judgments on
$S$, $V$, and $P$. Fine-grained agreement analysis for physical violation detection is provided in Appendix~\ref{sec:physical_validation}.
\section{Conclusion}
\label{sec:conclusion}
We introduced ReWorldBench for evaluating coding agents
on reconstruction, layout, and editing of realistic
3D scenes. By separating asset modeling from scene assembly,
the benchmark supports controlled layout tasks and
object-level physical validation.
Its complementary tasks and evaluation framework assess
how agents translate visual references into code, organize
objects in space, and modify scenes while preserving
unaffected content.
ReWorldBench provides a basis for assessing these capabilities
beyond successful execution and guiding the development
of agents that can reliably construct and edit 3D scenes.
\section*{LLM Usage Statement}
LLMs were used to assist with language polishing,
grammatical corrections, and improving manuscript clarity.
We also used VLMs and coding agents for dataset construction
and automated evaluation, including VLM-as-judge and
agent-as-judge methods, as described in the paper and
supplementary material.
All research ideas, experimental designs, analyses, and
conclusions were developed and verified by the authors.
The authors take full responsibility for the content
of this paper.
{
    \small
    \bibliographystyle{ieeenat_fullname}
    \bibliography{main}
}


\clearpage
\appendix
\section{Dataset Construction and Task Details}
\label{sec:supp_dataset}

\subsection{Data Construction Setup}
\label{sec:supp_construction_prompts}
For multi-agent visual preparation, we use GPT-6.1 Sol
(Codex CLI). Each agent operates in a separate session
with a designated role. Agents exchange JSON records,
code files, and image paths. A central controller coordinates
handoffs and tracks code versions, validation results,
and revision counts. The Object Analysis Agent and Scene
Planning Agent can invoke Gemini-3.1-Flash (Nano Banana 2)
to generate object-centric references and auxiliary scene
views, respectively.

We use GPT-6 Astra (Codex CLI) for three-stage code generation workflow and GPT-6.1 Sol (Codex CLI) to verify the corrections.
After scene acceptance, a camera agent using GPT-6.1 Sol
(Codex CLI) configures the eight predefined viewpoints
and renders the corresponding images.

\subsection{Memory Library}
\label{sec:supp_memory_mechanism}
We organize recurring modeling problems and their verified solutions by category and
store them as Markdown or JSON entries. After the generation agent applies a correction,
an independent validation agent checks whether it resolves the original problem and
whether it introduces any new issues. Human reviewers resolve ambiguous visual
judgments. The controller first stores verified feedback and corrections in a temporary
cache. After every five generation runs, it reviews the cache for recurring problems
that substantially affect visual fidelity or physical plausibility and have reusable
solutions. It consolidates overlapping records before adding them to the library.

In later runs of the three-stage generation workflow, the pipeline automatically selects
library methods before per-asset modeling in Stage 2. It uses the reference image and
asset requirements to assess each method's applicability and dependencies, then binds
the selected methods to the corresponding generation jobs and records their versions and
parameters. Assets without a suitable entry follow the standard generation procedure.
For the rooftop garden, for example, the workflow first builds the sofa seat and
backrest, then invokes the selected support-fitting method to generate five cushions.
After assembly, it checks the execution records to verify that all five cushions are
included in the final scene.

\begin{figure}[htbp]
    \centering
    \includegraphics[width=\linewidth]{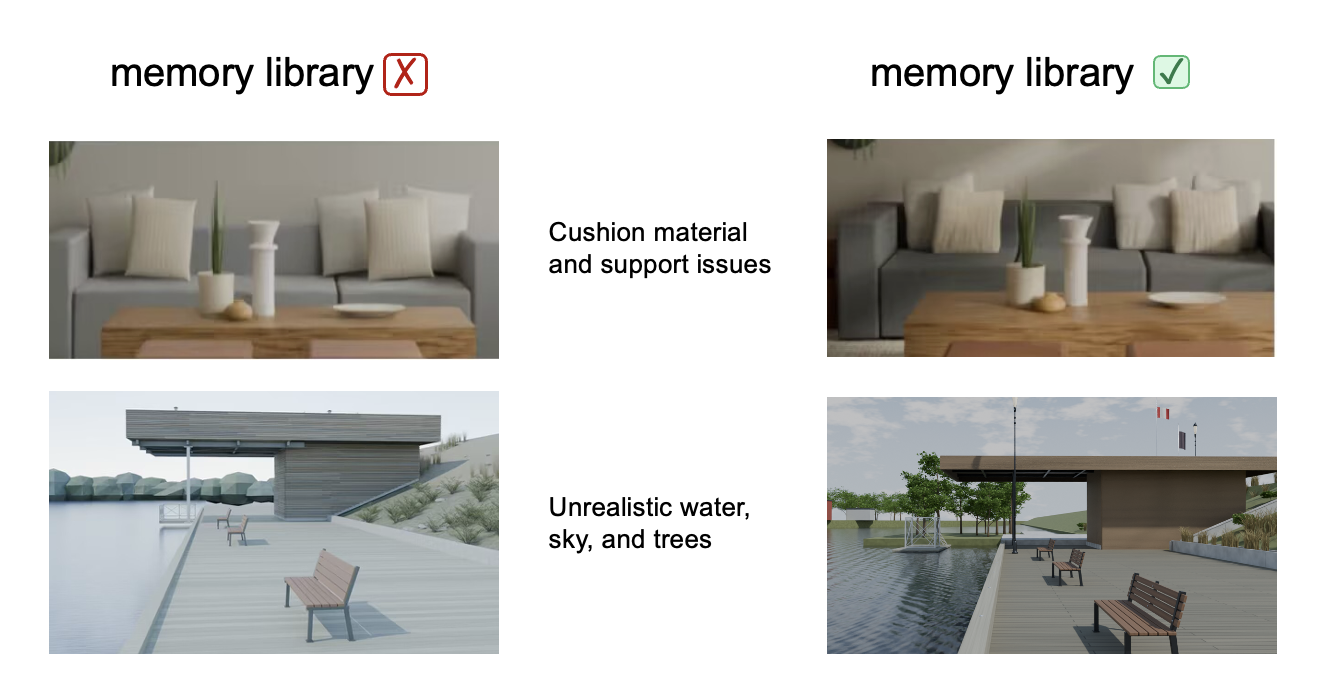}
    \caption{\textbf{Two types of modeling problems addressed by the memory library
    during dataset construction.} The left and right columns show the initial and
    revised results, respectively.}
    \label{fig:memory}
    \vspace{-10pt}
\end{figure}

Figure~\ref{fig:memory} shows how revisions address two types of modeling problems. In the indoor
scene, the initial cushions have regular outlines, limited variation in fabric
appearance and folds, and a stiff pose against the sofa. The revised cushions have more
natural contours and folds, more varied shading across the fabric, and a more convincing
fit against the sofa. In the waterfront scene, the initial water surface is smooth and
uniform, and the sky and trees contain little detail. The revised water shows fine
ripples and clearer reflections, while the sky and vegetation add depth and make the
water, walkway, and background easier to distinguish.

\subsection{Quality Control}
\label{sec:supp_scene_acceptance}

\noindent\textbf{Scene acceptance.}\quad
We allow at most three revision rounds per scene.
After each revision, we execute the updated program and
reassess its physical and visual quality. The controller
retains a scene only when the same code version achieves
a physical score of 100 and a visual score above 80.
The staged workflow and independent asset implementations
help the generation agent trace validation feedback to
the relevant code and make targeted corrections.
Scenes that fail either criterion after the revision limit
are excluded. 

\noindent\textbf{View configuration.}\quad
We specify eight viewpoints: front, rear, left, right,
top-down, front-left, front-right, and elevated front.
For each accepted scene, our camera agent uses GPT-6.1 Sol (Codex CLI) to set the camera pose for each viewpoint and render the corresponding image.
All eight views therefore share the same underlying
scene geometry and asset arrangement.

\noindent\textbf{Reference-image verification.}\quad
We use Gemini-3.1-Flash (Nano Banana 2) to refine surface
textures under explicit preservation constraints.
We provide the original render together with the prompt below.

Human reviewers compare each refined image with its
original render and check consistency across views.
They assess whether object identities, instance counts,
geometry, and spatial relationships are preserved.
We retain only images that satisfy these requirements.
Reviewers reject images that alter scene content or
structure, even if their surface appearance improves.

\begin{tcolorbox}[
    enhanced,
    unbreakable,
    colback=white,
    colframe=gray,
    boxrule=0.6pt,
    arc=2mm,
    left=3mm,
    right=3mm,
    top=5mm,
    bottom=3mm,
    before skip=12pt,
    after skip=12pt,
    fontupper=\small,
    fonttitle=\small\sffamily\bfseries,
    coltitle=black,
    title={Image Refinement Prompt},
    attach boxed title to top center={yshift=-2mm},
    boxed title style={
        colback=gray!25,
        colframe=gray!25,
        boxrule=0pt,
        arc=1.5mm,
        left=2mm,
        right=2mm,
        top=1mm,
        bottom=1mm
    }
]
You are an image refinement agent. Your task is to refine
only the surface textures in the input Blender render while
preserving the scene exactly as depicted.

Enhance fine details such as material grain and subtle texture
variation. Preserve existing material types, dominant colors,
and recognizable patterns. Do not add seams, folds, panels,
or protrusions.

Keep the camera viewpoint, perspective, framing, lighting,
shadows, and image dimensions fixed. Preserve every object's
identity, shape, size, position, and orientation. Maintain
architectural boundaries, contact points, support relationships,
and occlusion order. Do not add, remove, duplicate, or replace
objects, or reconstruct hidden parts.

If a texture edit could alter an object's apparent shape or
its contact with another surface, leave that region unchanged.
Return only the refined image.
\end{tcolorbox}

\begin{figure*}[t]
\centering
\newcommand{\datasetcard}[3]{%
    \begin{minipage}[t]{0.242\textwidth}
    \begin{tcolorbox}[
        width=\linewidth,
        colback=white,
        colframe=gray!50,
        boxrule=0.4pt,
        arc=1mm,
        boxsep=0pt,
        left=0.6mm,
        right=0.6mm,
        top=0.6mm,
        bottom=0.8mm,
        before skip=0pt,
        after skip=0pt
    ]
        \centering
        \setlength{\parskip}{0pt}%
        \parbox[c][30mm][c]{\linewidth}{%
            \centering
            \includegraphics[
                width=\linewidth,
                height=30mm,
                keepaspectratio
            ]{fig/appendix/datasets/#1}%
        }\par\nointerlineskip
        \parbox[c][5.5mm][c]{\linewidth}{%
            \centering\scriptsize\bfseries #2%
        }\par\nointerlineskip
        \parbox[c][3.5mm][c]{\linewidth}{%
            \centering\fontsize{6}{7}\selectfont Category: #3%
        }\par
    \end{tcolorbox}
    \end{minipage}%
}

\datasetcard{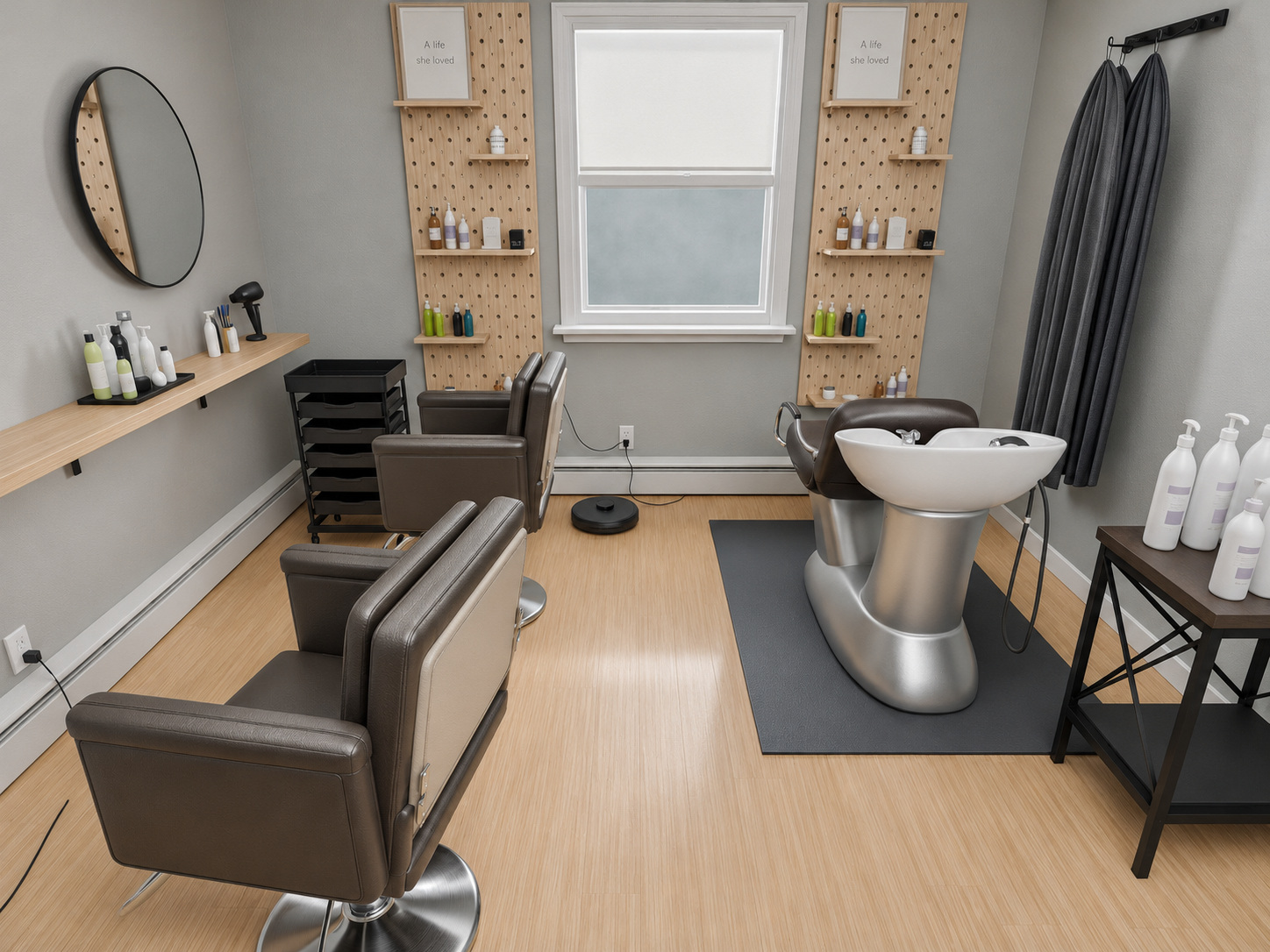}
    {Barber Shop}{Personal Care}%
\hfill
\datasetcard{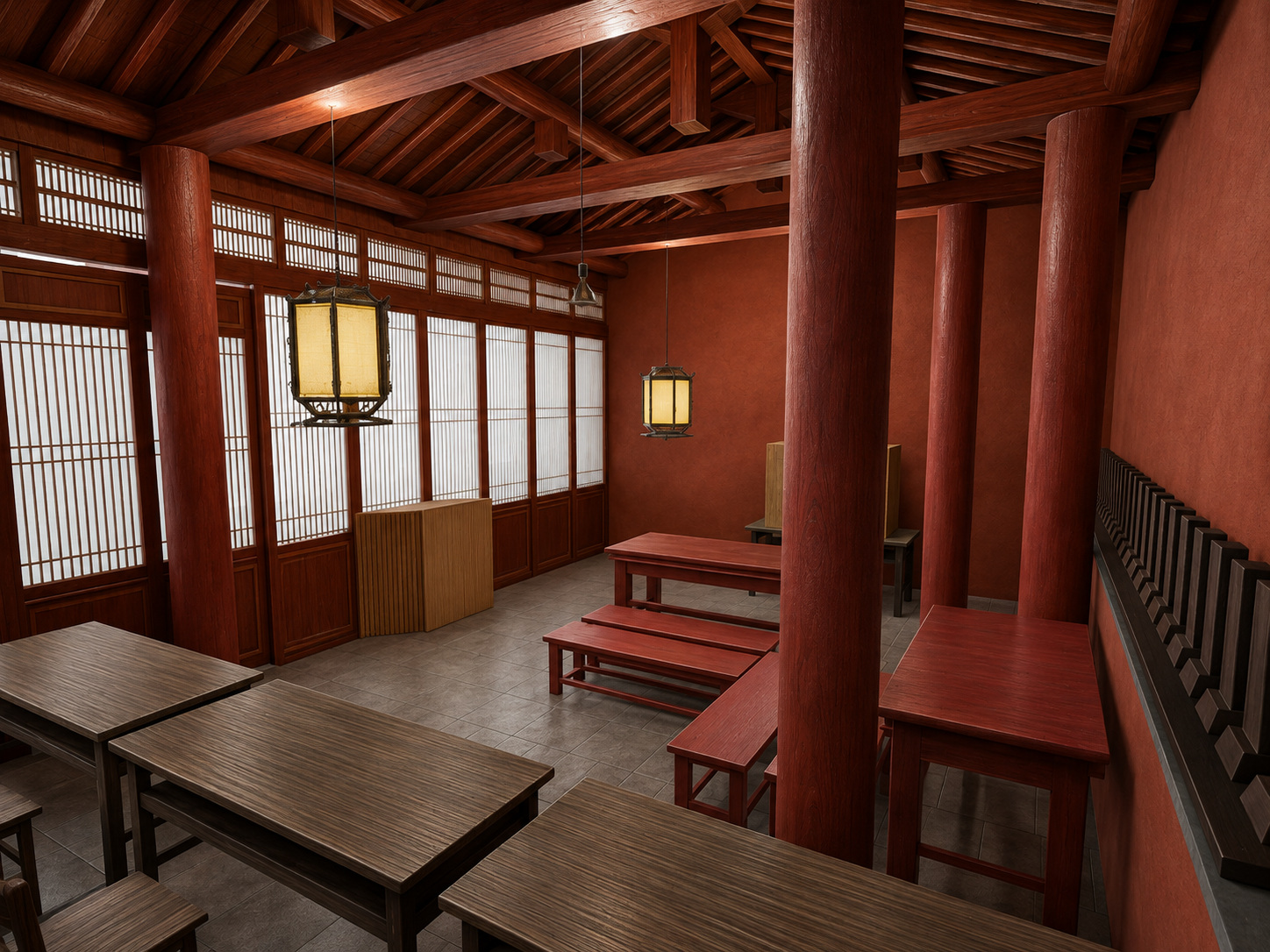}
    {Traditional Ceremonial Hall}{Traditional Hall}%
\hfill
\datasetcard{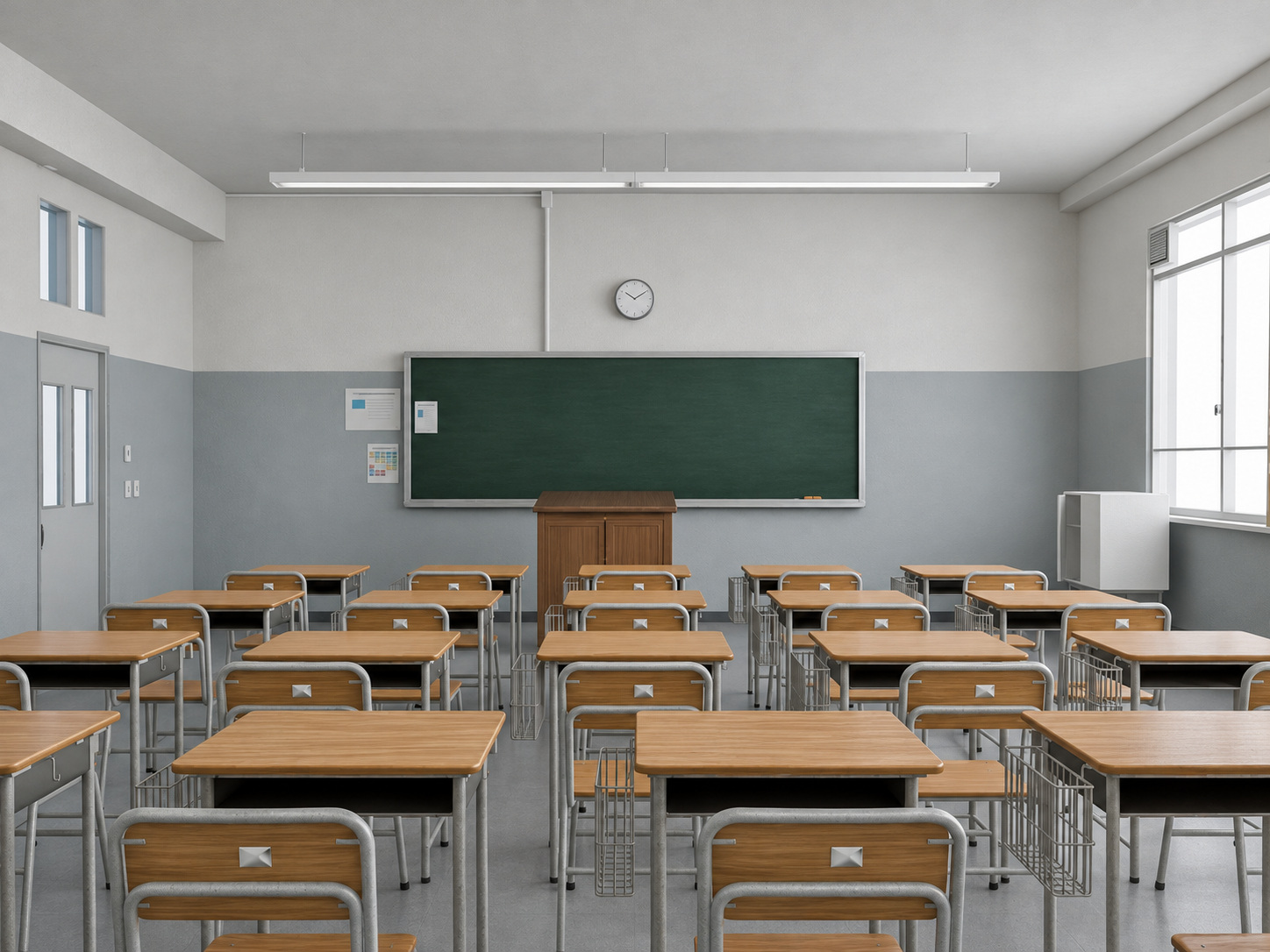}
    {Classroom}{Education and Reading}%
\hfill
\datasetcard{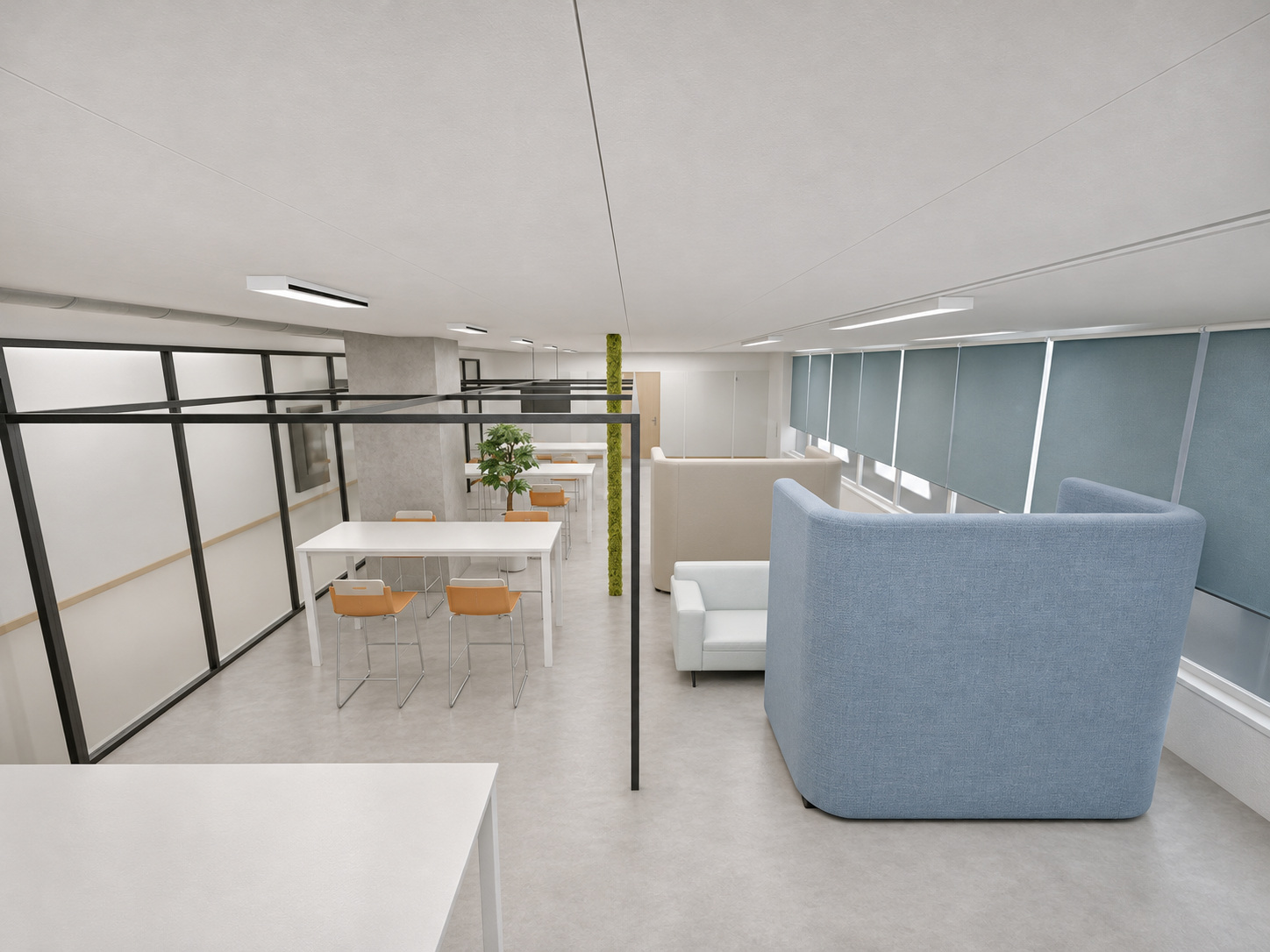}
    {Collaborative Office}{Office}

\par\vspace{1.2mm}

\datasetcard{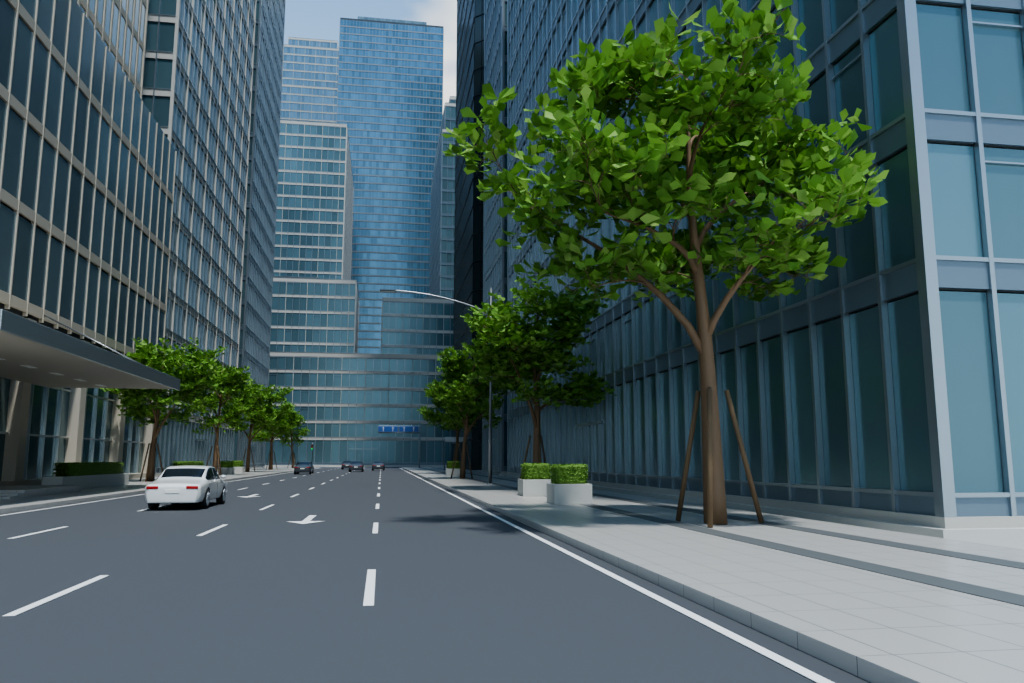}
    {Urban Street}{Streetscapes}%
\hfill
\datasetcard{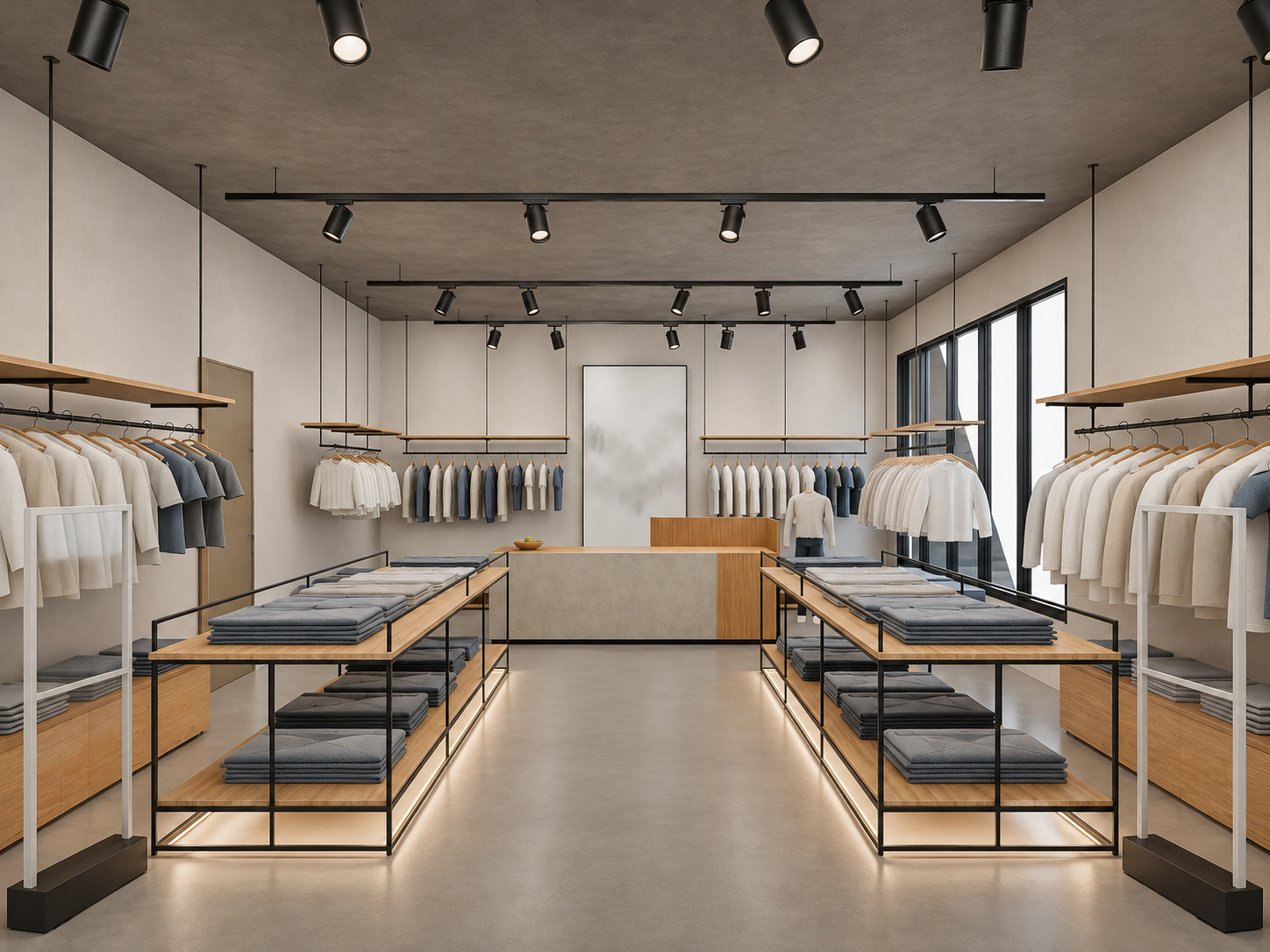}
    {Clothing Store}{Retail}%
\hfill
\datasetcard{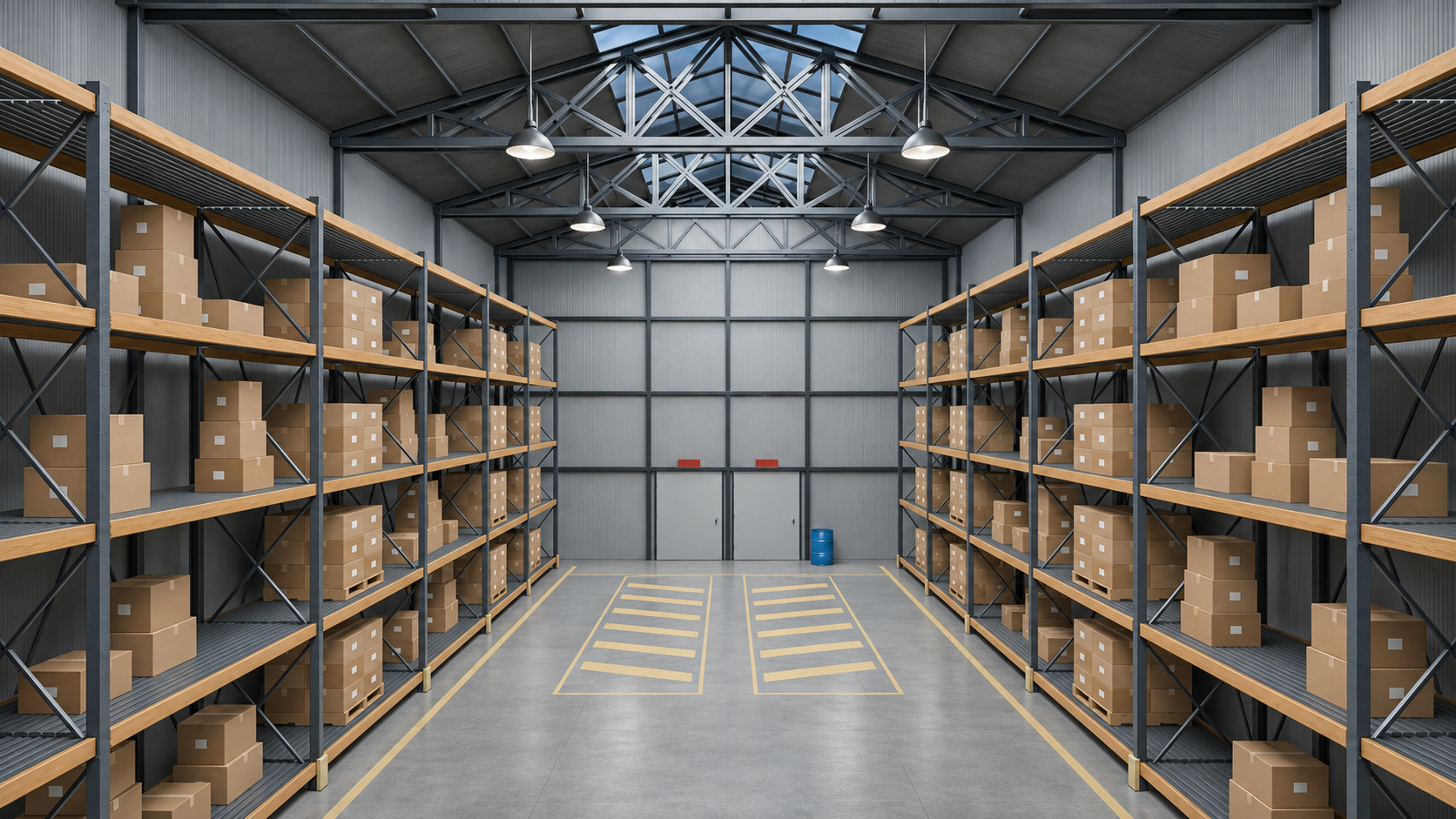}
    {Warehouse}{Storage and Logistics}%
\hfill
\datasetcard{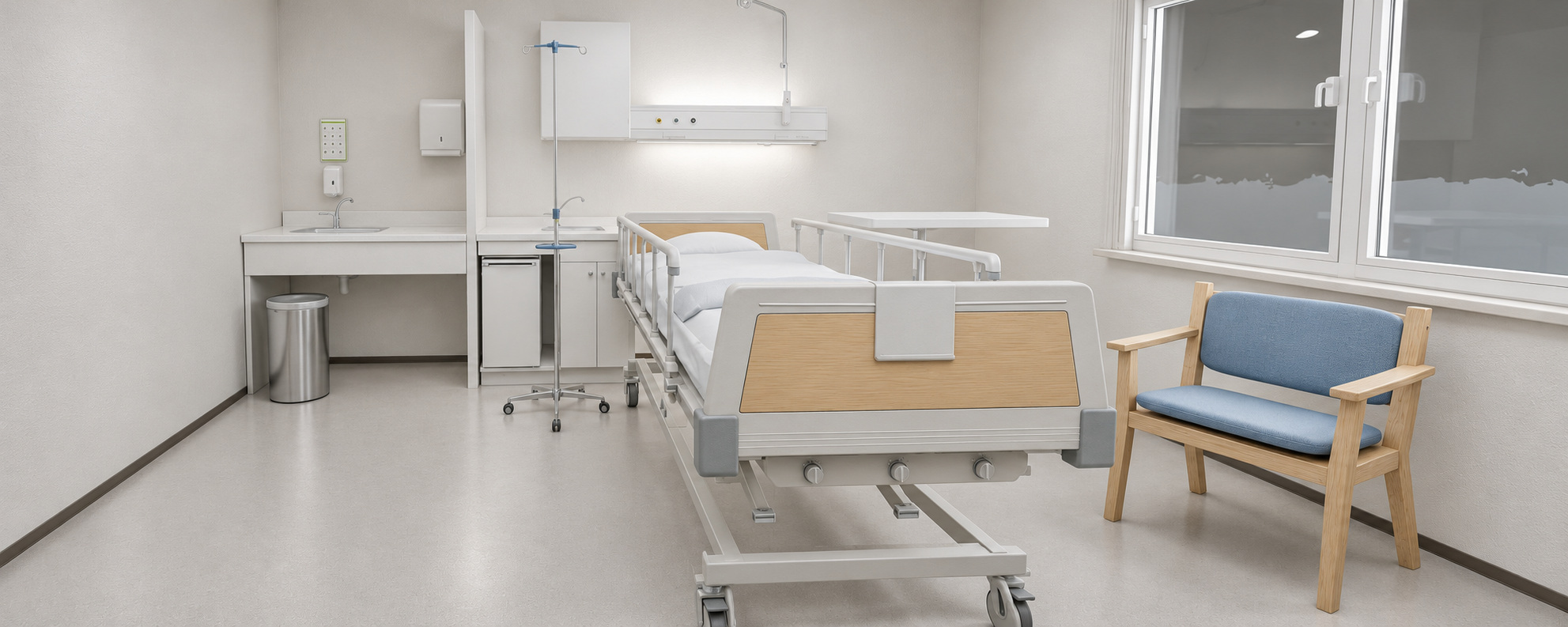}
    {Hospital Ward}{Healthcare}

\par\vspace{1.2mm}

\datasetcard{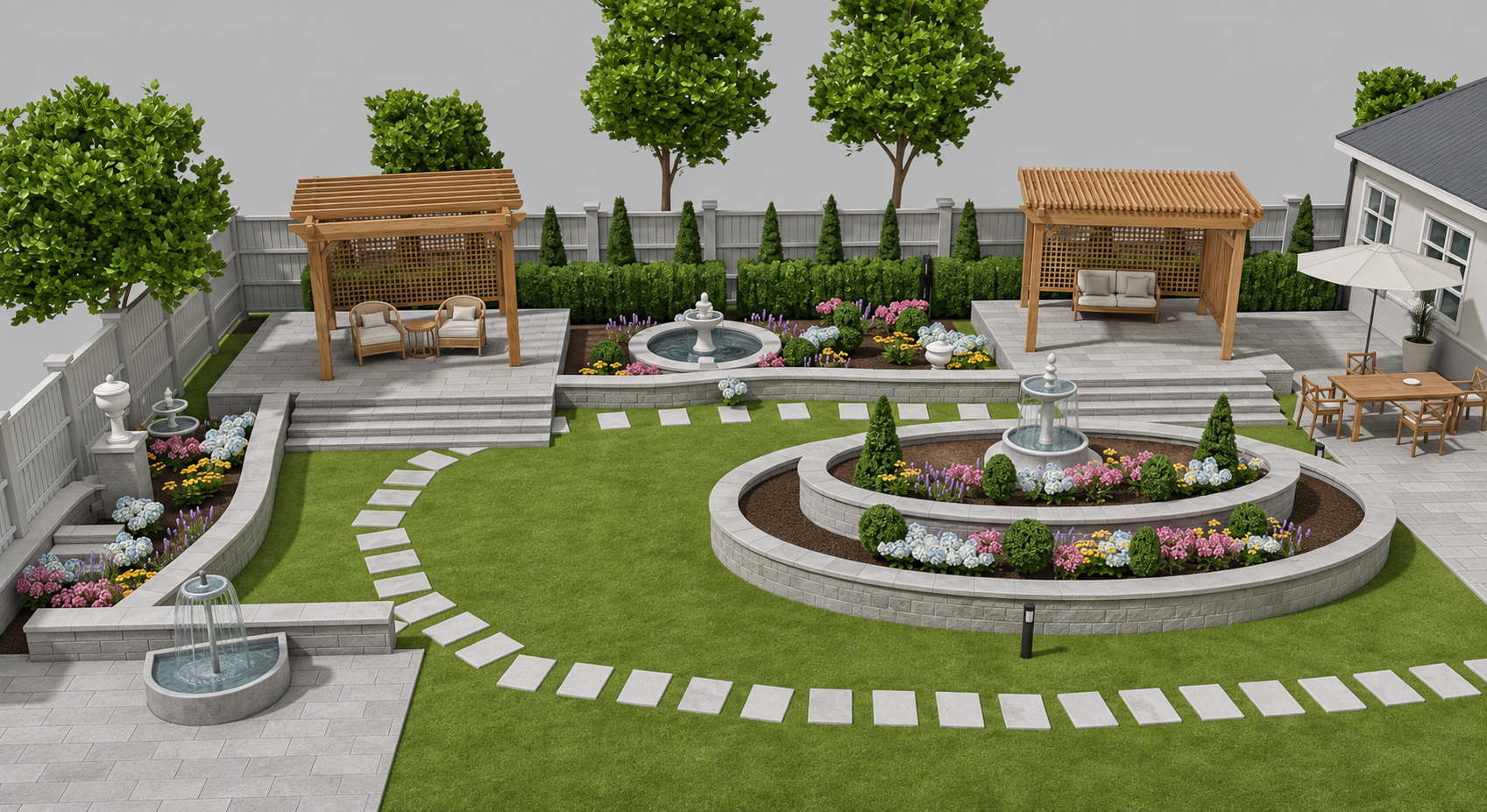}
    {Landscaped Garden}{Park and Plaza}%
\hfill
\datasetcard{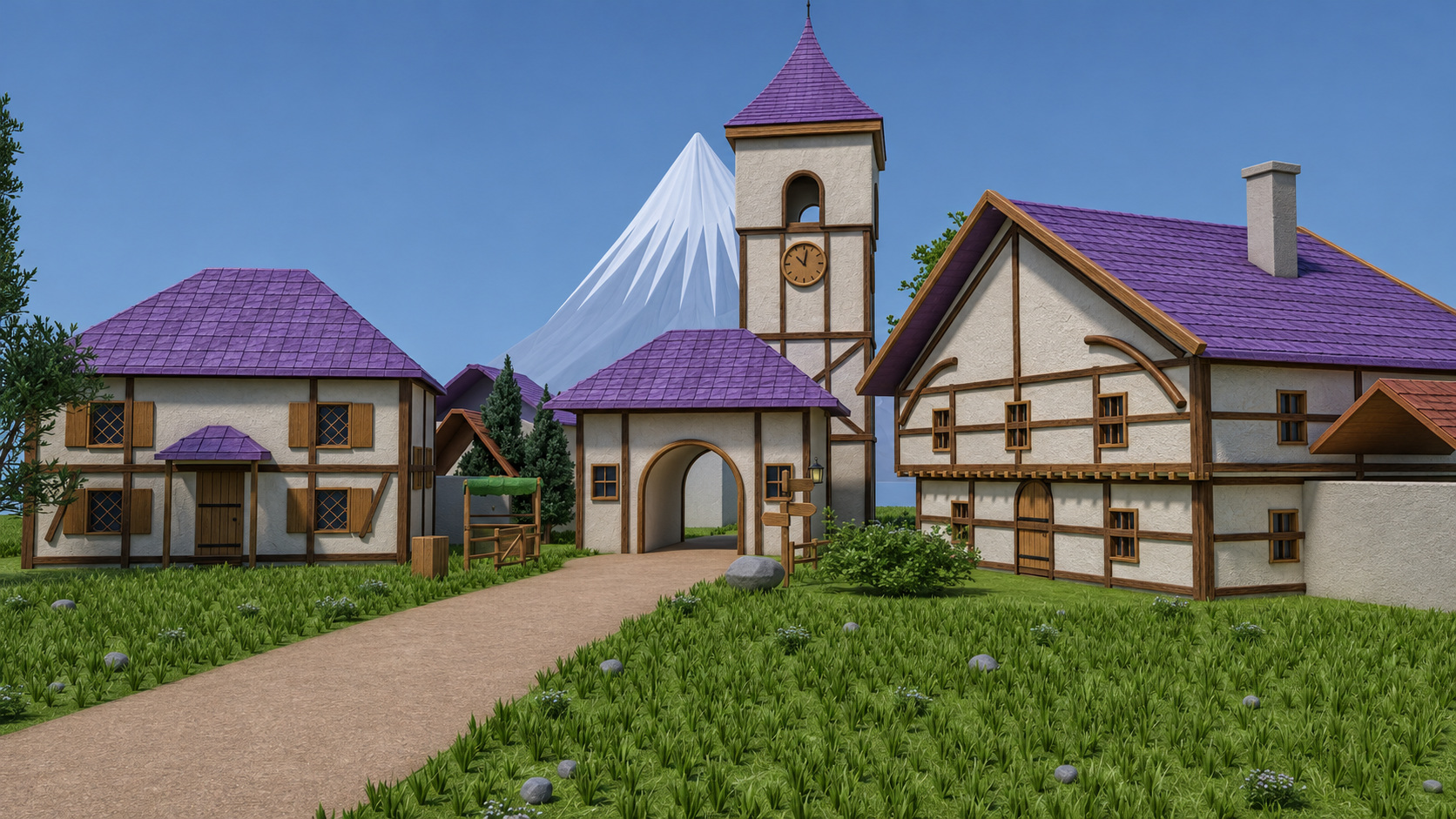}
    {Fantasy Village}{Street and Townscape}%
\hfill
\datasetcard{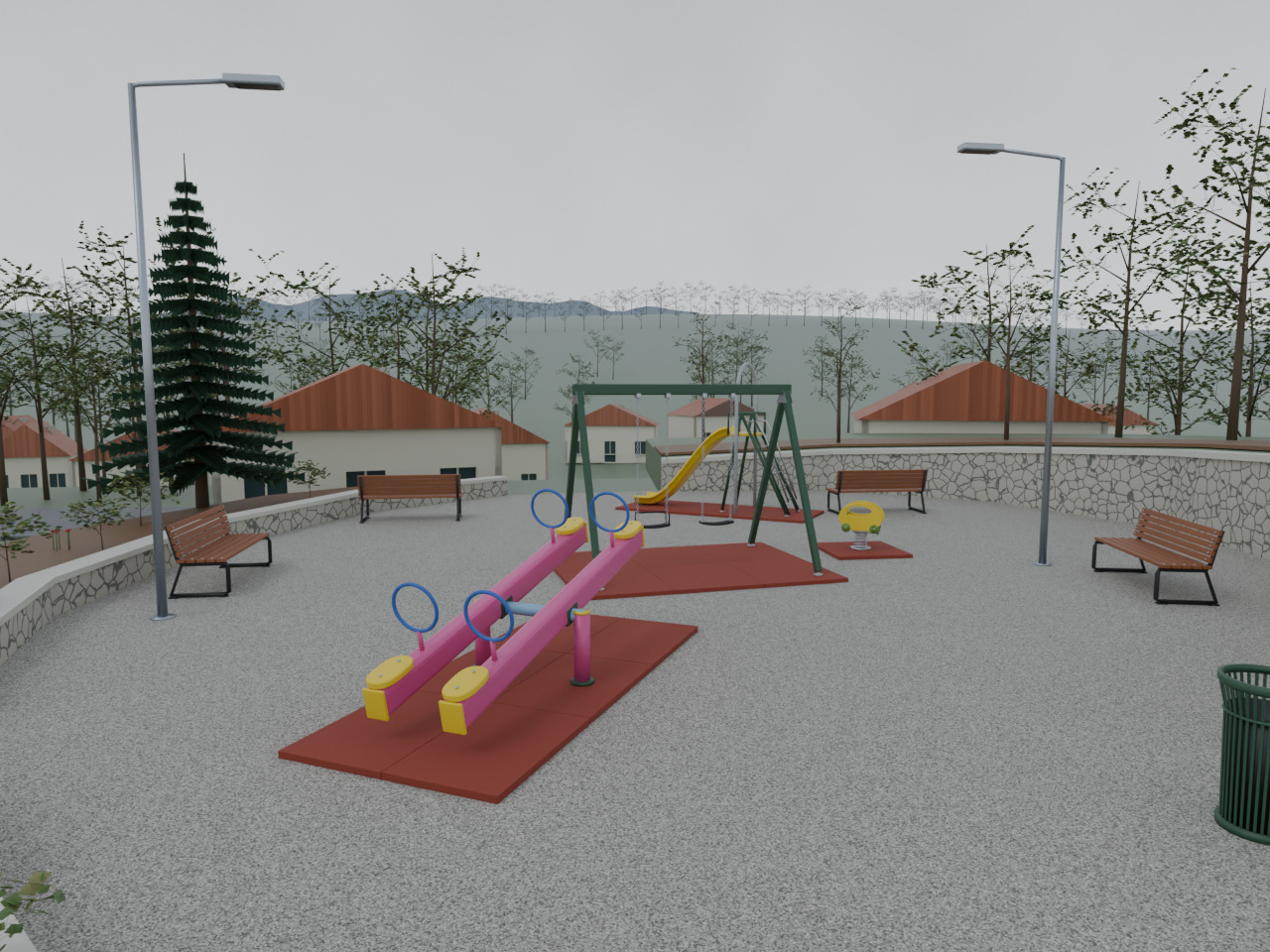}
    {Playground}{Sports and Playground}%
\hfill
\datasetcard{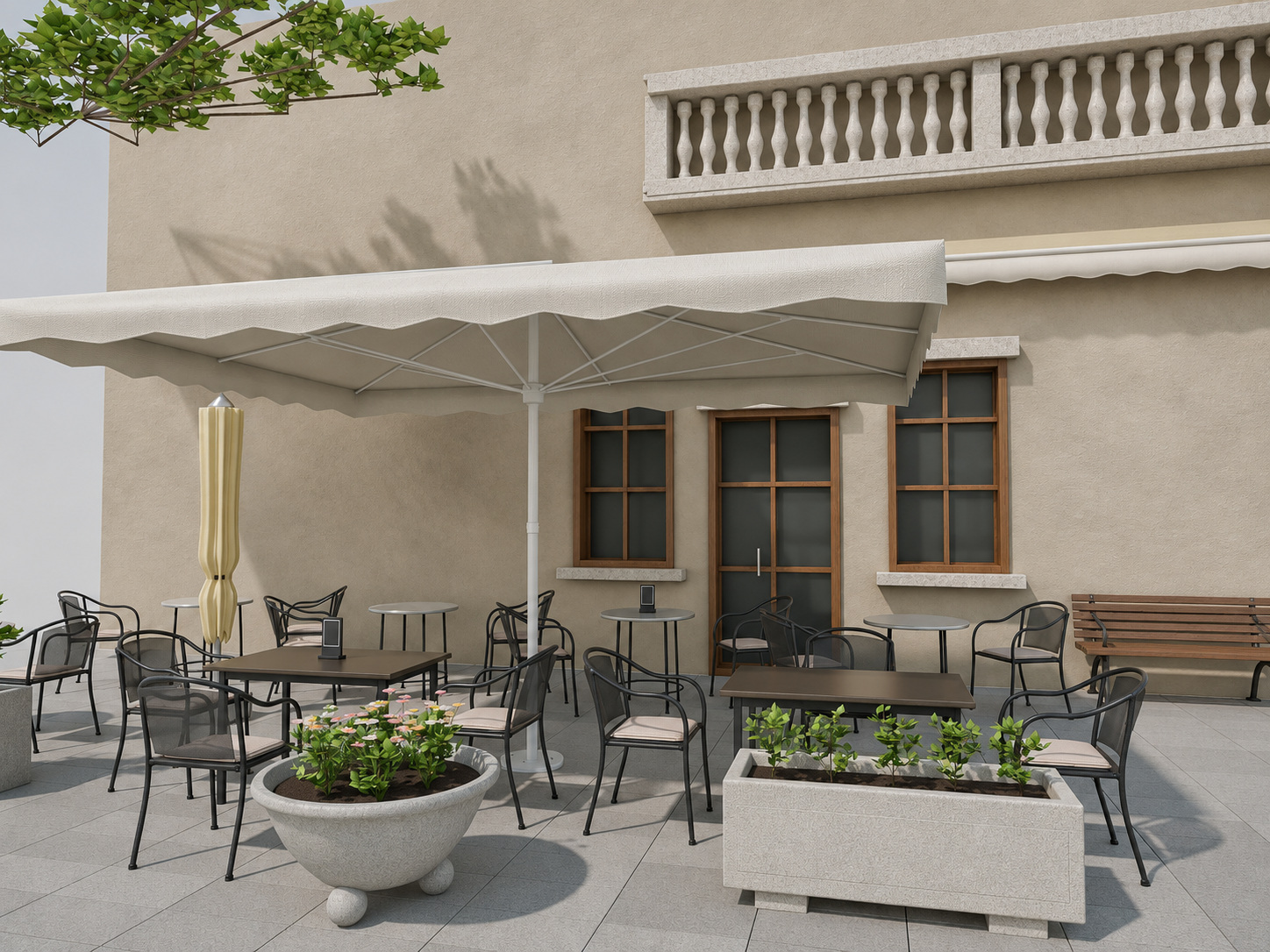}
    {Outdoor Restaurant}{Outdoor Dining}

\par\vspace{1.2mm}

\datasetcard{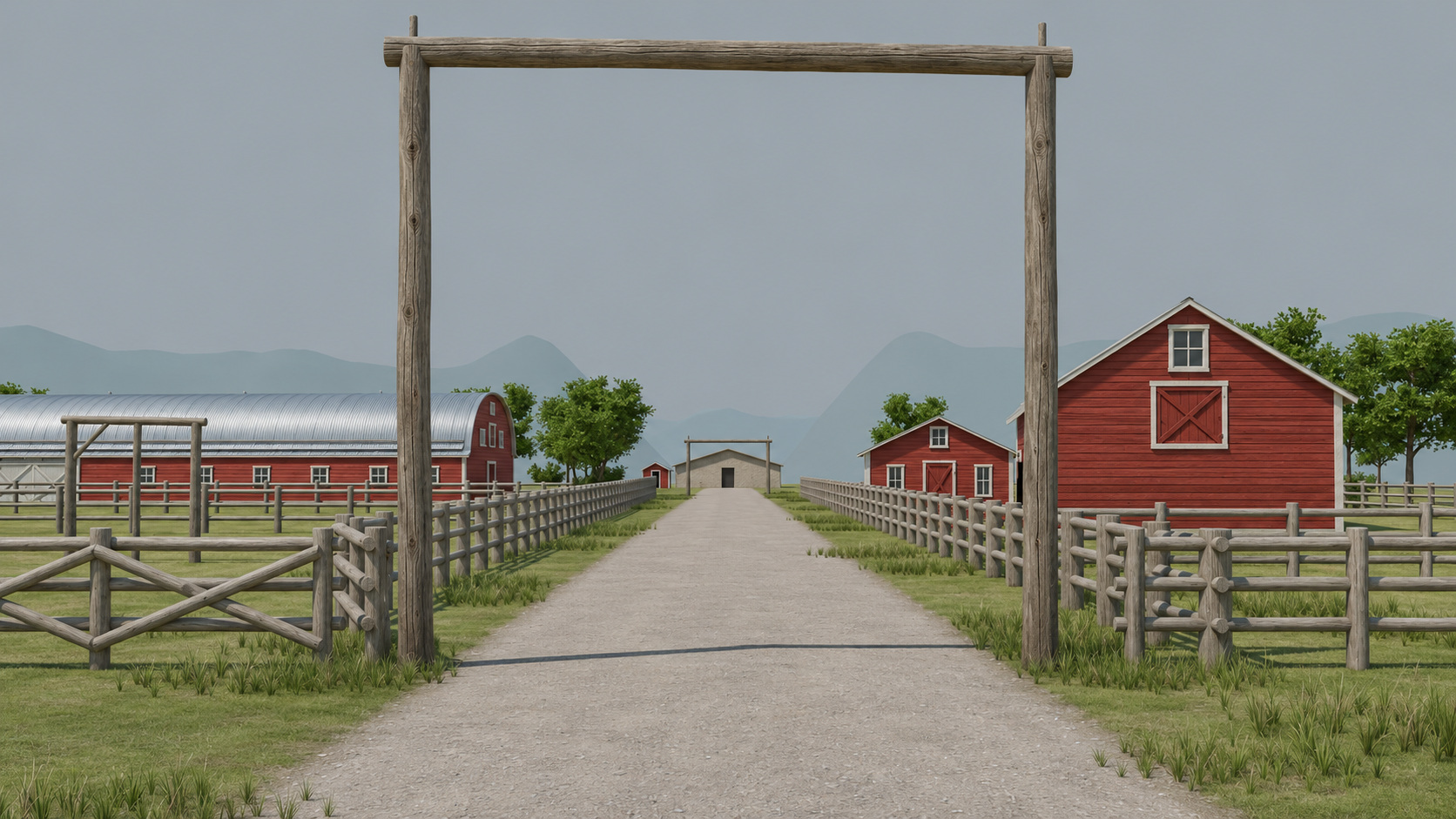}
    {Farm}{Agricultural}%
\hfill
\datasetcard{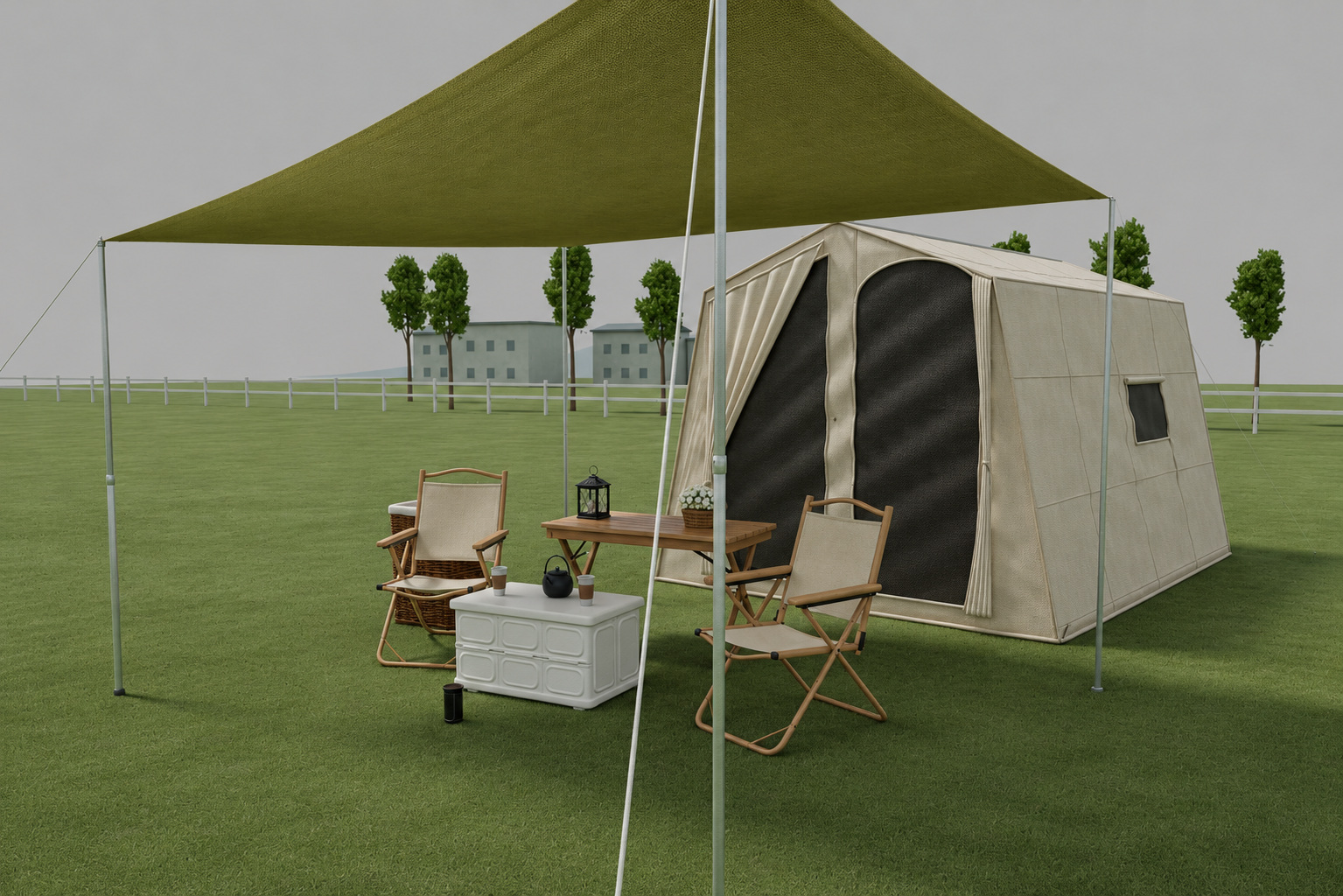}
    {Grassland Camping}{Terrace and Camping}%
\hfill
\datasetcard{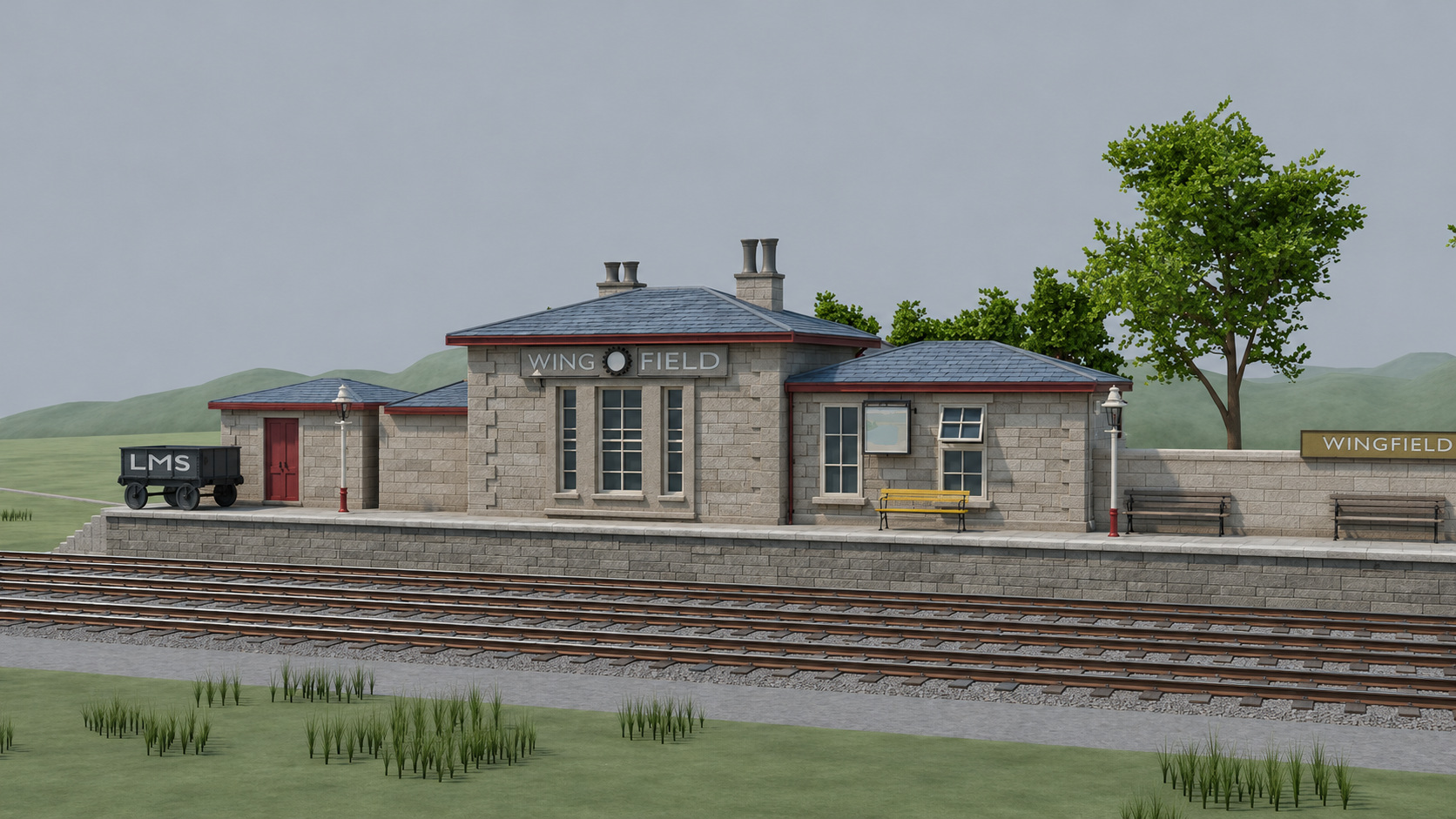}
    {Train Station}{Transportation}%
\hfill
\datasetcard{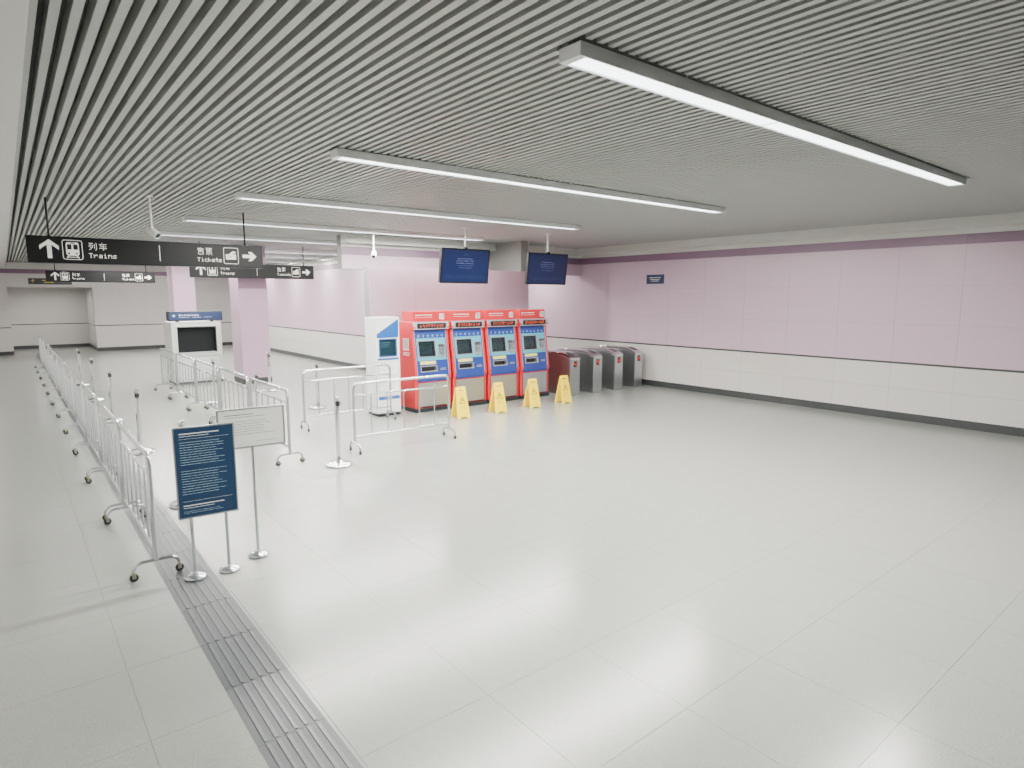}
    {Subway Station}{Transportation}

\caption{Representative indoor and outdoor scenes with their categories.
The playground image is an original Blender render; the remaining
images are task reference images.}
\label{fig:scene_acceptance_example}
\vspace{-10pt}
\end{figure*}

Figure~\ref{fig:scene_acceptance_example} presents
representative scenes from our dataset, covering
a range of indoor and outdoor categories.

\subsection{Task Construction}

\noindent\textbf{Image-conditioned reconstruction.}\quad
For each scene, we select reference images from eight rendered views. We start with the front view and manually check whether every object is visible and identifiable. If any object is missing from view or too occluded to identify, we add views that reveal it. We aim to use as few views as possible while ensuring that each object can be identified in at least one view.

We construct the task prompt by combining a shared instruction with a scene-specific description. The shared instruction asks the agent to reconstruct the scene from the reference images as executable Blender code. To generate the scene-specific description, a model analyzes the selected views and describes the objects and the scene’s visual style.

\noindent\textbf{Asset-constrained layout.}\quad
We extract the architectural structure code and individual
asset code from the modeling stage and provide them as
the task inputs, together with the target reference image.
We use a shared prompt that asks the agent to position
and orient the supplied assets to match the reference,
while preserving asset geometry and the architectural
structure.

\noindent\textbf{Instruction-guided editing.}\quad
We use existing Blender scenes as starting points and provide the agent
with the initial scene code, rendered views, and a scene-specific editing
instruction (A). We write each instruction around the target objects and the
required edits or spatial constraints. For fault-repair tasks, we introduce
geometric defects into the initial scene. For infeasibility judgment tasks,
we specify constraints that cannot be satisfied simultaneously. Each
instruction is paired with shared acceptance requirements (B1) and
task-specific criteria (B2).
Appendix~\ref{sec:editing_task_examples_figure} provides examples of all six
editing settings.

\section{Implementation and Experimental Settings}
\label{sec:supp_implementation}

\subsection{Models and Agent Configurations}
\label{sec:supp_models}
Among the eight models evaluated in ReWorldBench, we run
Claude Opus 5.5 through Factory CLI and the other seven
through Codex CLI. We use the \texttt{high} reasoning setting
for all agents and allow each agent 3,600 seconds per task
instance.

We provide the task inputs and instructions and allow
each agent to analyze them autonomously. For tasks
requiring code, the agent must submit a final code file
within the allotted budget. If the agent reaches the
time limit without submitting the required file, we
record an agent timeout and mark the task as failed.
The infeasibility-judgment setting (T4) instead permits
a refusal, which is evaluated according to the
task-specific scoring rules.


\subsection{Input-View and Feedback Experiment Details}
This supplementary experiment uses the same generation budget as the main benchmark, with 3,600 seconds each for execution and rendering. The comparison uses the same 20 scenes (12 indoor and eight outdoor) and six conditions, yielding 120 scored outputs. Three initially failed generation runs were retried before scoring. Some scenes use compatible Compact S/Q rubric mappings and have limited reference-view coverage; within each scene, all conditions use consistent criteria. Feedback starts from the four-view output, and the reported final change uses the third round rather than the best round. The 1.55-point gain is computed from unrounded paired scores.

\section{Additional Results and Visual Examples}
\label{sec:supp_additional_results}

\subsection{Additional Qualitative Results}
\label{sec:additional_qualitative}

Figure~\ref{fig:qualcase_reconstruction} compares reconstruction
results for a garden and a ceremonial hall.
In the garden, GPT-6 Astra, GPT-6.1 Sol, and Kimi-K3.1-Pro
produce more detailed foliage and pergola structures.
Claude Opus 5.5 and Qwen3.8-Max use smoother, more rounded
tree crowns, while DeepSeek-V4.1-Flash exhibits visibly
faceted vegetation.

In the ceremonial hall, GPT-6 Astra exhibits visible
interpenetration between furniture components.
Qwen3.8-Max and DeepSeek-V4.1-Flash deviate from the
reference table placement and spacing.
Claude Opus 5.5, DeepSeek-V4.1-Flash, and GLM-5.3-Flash
also simplify or alter the reference arrangement of
ceiling beams and rafters. These cases reveal differences
in geometric detail, layout fidelity, and physical
consistency among otherwise recognizable reconstructions.

Figure~\ref{fig:task2_qualitative} shows that placement
errors persist in the asset-constrained layout task.
Despite receiving predefined assets, models produce
misoriented chairs, interpenetrating objects, and
in some cases omit core scene elements.
Together, these examples highlight challenges in both
modeling individual components and arranging them
into a complete, physically consistent scene.

Figures~\ref{fig:task2_qualitative}--\ref{fig:qualcase_autonomous} compare the visual outcomes of instruction-guided editing. In the farm scene, several outputs turn the barn dark green while retaining its white trim, whereas others leave the barn red. In the clothing store and library, outputs differ in how the relocated checkout counter aligns with the display fixtures and where the statues are placed relative to the relocated bookshelves. In the classroom, some outputs lower the raised desk near the window, while others leave it elevated. The rooftop case shows both correct refusals and attempted rearrangements for an infeasible request. On the restaurant terrace, both the final position of the bench and the arrangement of nearby seating vary across outputs. These results reveal a gap between local object manipulation and scene-level spatial reasoning. While models can perform explicit attribute changes, their ability to maintain spatial relationships, satisfy inter-object constraints, and recognize infeasible requests remains inconsistent.

\subsection{Agent Trajectory Analysis under Visual Feedback}
\label{sec:feedback_trajectories}

\begin{figure*}[t]
    \centering
    \includegraphics[width=\textwidth]{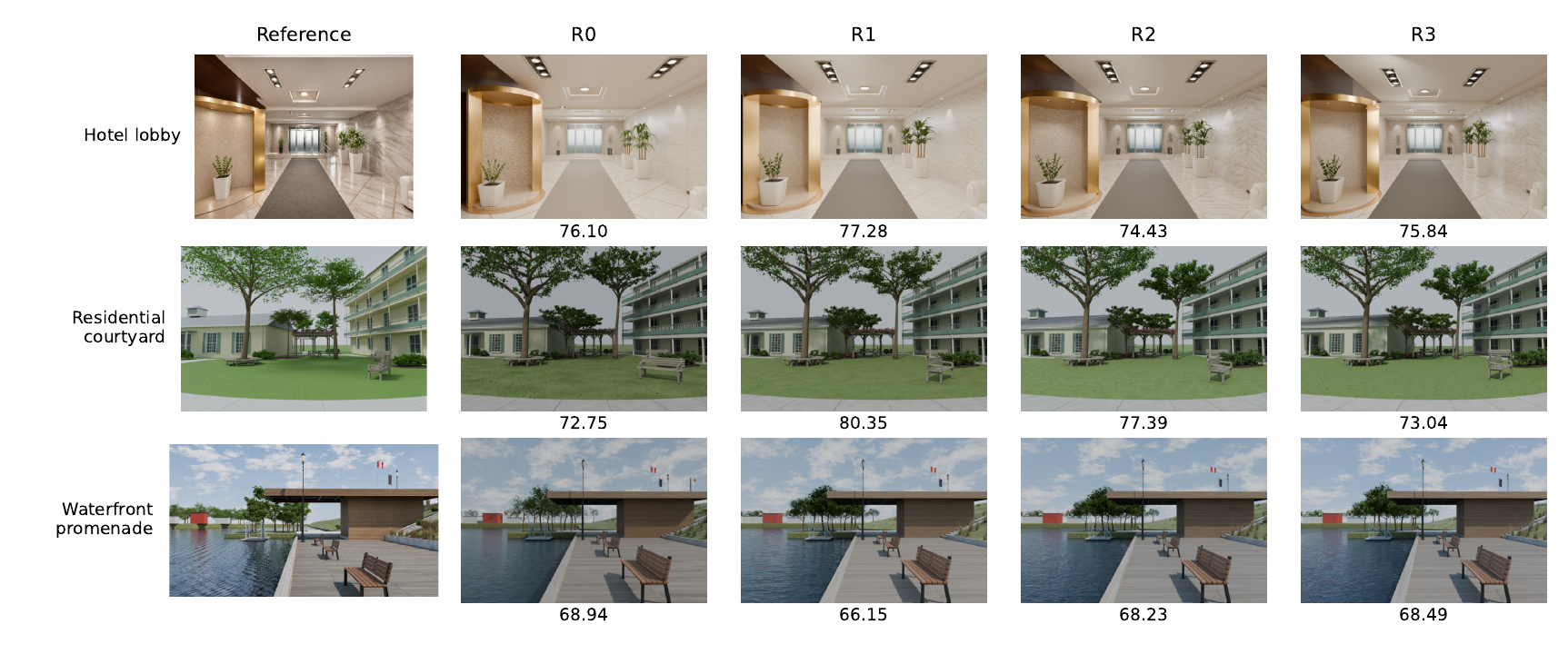}
    \caption{\textbf{Reconstruction trajectories under visual feedback.} Each row shows
    one reference image, the four-view baseline (R0), and three successive revisions
    (R1--R3). The four renders in each row use the same main evaluation camera. Numbers
    below the renders indicate overall scores on a 0--100 scale.}
    \label{fig:feedback_trajectories}
    \vspace{-10pt}
\end{figure*}

Figure~\ref{fig:feedback_trajectories} illustrates the revision process for three scenes
under visual feedback. In each round, the agent compared the reference images with the
previous renders, identified visual discrepancies, and revised the code. The figure
shows the four-view baseline (R0) followed by three feedback rounds (R1--R3). The
numbers below the renders are overall scores.

\noindent\textbf{Hotel lobby.}\quad
The agent identified faint marble veining, weak floor reflections, and a carpet that was
lighter than in the reference. In R1, it darkened the carpet and increased the
visibility of the marble veining and floor reflections. Later rounds revised the stone,
brass finishes, and lighting. By R3, reference similarity and visual quality had
improved over the baseline, but physical validation deducted points for insufficient
floor support of major furniture and fixtures. The overall score changed from 76.10 to
75.84.

\noindent\textbf{Residential courtyard.}\quad
The agent identified a mismatch in the orientation of the foreground bench. In R1, it
set the bench angle to $34^{\circ}$ and revised the railings and vegetation, raising the
overall score from 72.75 to 80.35. In R2, it adjusted the bench angle to $39^{\circ}$
and revised the tree crowns and paving, while reference similarity fell from its R1
level. In R3, it reduced the lawn extent and moved the bench. Object intersections again
incurred a deduction, and the final score was 73.04, only 0.29 points above the
baseline.

\noindent\textbf{Waterfront promenade.}\quad
In R1, the agent identified an abrupt cutoff at the foreground water boundary. It
extended the water surface and adjusted reflections, vegetation, and lighting under the
pavilion. In R2, it flagged ripples that were too wide and overly parallel, then refined
their pattern. By R3, visual quality had risen from 50.59 to 58.45, while reference
similarity had fallen from 76.25 to 67.65. The overall score changed from 68.94 to
68.49.

These examples show that the agent can revise scenes in response to visual feedback, but
changes to appearance and layout may lead to physical issues such as object
intersections or insufficient support, offsetting visual gains. Multiple feedback rounds
therefore do not consistently improve overall quality, indicating that the agent remains
limited in coordinating visual and physical requirements during refinement.

\subsection{Human Agreement on Physical Scoring}
\label{sec:physical_validation}
\begin{figure}[H]
    \centering
    \includegraphics[width=\linewidth]
        {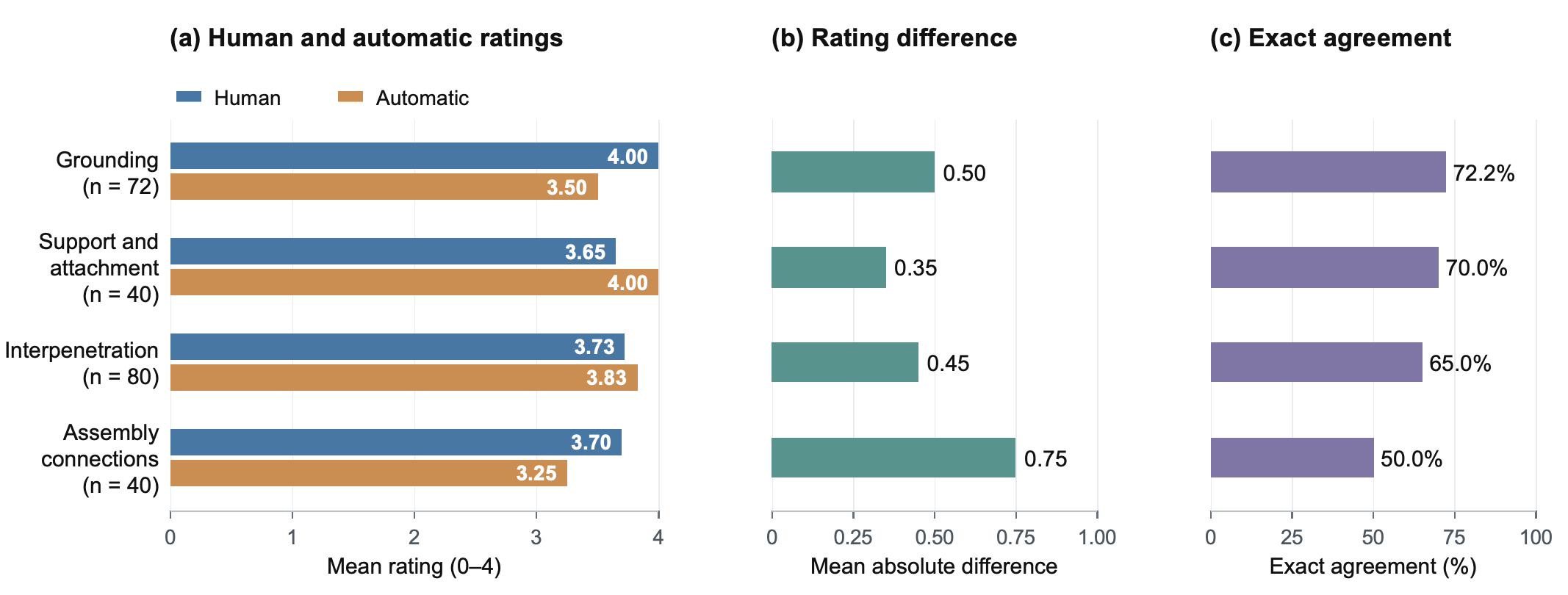}
    \caption{Human--agent agreement on physical scoring
    across 232 paired rubric items from 40 reconstruction
    outputs. (a) Mean ratings on the 0--4 scale.
    (b) Mean absolute rating differences.
    (c) Exact agreement rates.
    Numbers in parentheses indicate paired item counts.}
    \label{fig:physical_category_agreement}
    \vspace{-10pt}
\end{figure}

We compare automatic physical scores with human ratings
on 40 randomly sampled reconstruction outputs.
We first run the agent judge, which assesses each output
and collects additional evidence, including targeted
renders for specific rubric items.
A human evaluator then examines the four standard views
and can consult the same supplementary evidence when
rating the corresponding items.
Both evaluators apply the same scene-specific rubrics
and have access to the same collected evidence.
We organize the comparison into four categories:
grounding, support and attachment, interpenetration,
and assembly connections. All ratings use the original
0--4 scale.

Figure~\ref{fig:physical_category_agreement} summarizes
the results. Human and automatic ratings average 3.80
and 3.66, respectively, giving a mean signed difference
of 0.14 points. The mean absolute difference across
paired items is 0.50 points, and 65.5\% of ratings
match exactly. Thus, the two evaluators provide similar
average scores, although individual judgments differ.

Agreement varies across physical relations.
Support and attachment has the smallest mean absolute
difference (0.35), while grounding has the highest
exact agreement (72.2\%).
Assembly connections shows the largest discrepancy,
with a mean absolute difference of 0.75 and an exact
agreement rate of 50.0\%.
Human ratings are higher on average for grounding
and assembly connections, but lower for support
and attachment and interpenetration.

\section{Physical Validation Details}
\label{sec:supp_physical_validation}

Both reconstruction and layout tasks use Blender-based
geometric checks to assess physical plausibility.
Their validation procedures differ in how objects and
relationships are identified and how the resulting
evidence is assessed.

\noindent\textbf{Agent-based validation for reconstruction.}\quad
Reconstructed scenes may use arbitrary mesh decompositions
and object names, so their components cannot be directly
matched to predefined asset identities. Guided by a
scene-specific rubric, an evaluation agent identifies
semantic objects from the rendered views and scene
inventory, then examines the required physical relationships.
It first locates apparent violations or uncertain interfaces.
The agent then performs targeted mesh checks and, when
necessary, requests additional renders from viewpoints
that expose the relevant contact or intersection.

The agent combines geometric measurements with visual
evidence to assess each relationship under the rubric's
severity and extent criteria. Each finding records the
affected objects, relation, and supporting evidence.
Concealed interfaces and conflicting observations are
reported with explicit limitations and reduced confidence.

\noindent\textbf{Object-level validation for layout.}\quad
Layout tasks provide the asset and architectural code,
which establishes the identities and component membership
of the supplied assets. We use these correspondences to
group candidate meshes and evaluate the specified
relationships directly in Blender. The validator applies
fixed geometric rules and tolerances to the resulting
object groups. This enables repeatable object-level checks
while avoiding the object-identification and iterative
evidence-acquisition steps required for reconstruction.

\subsection{Validation Criteria}
\label{sec:supp_validation_criteria}

The two procedures draw on the geometric checks described
below. The reconstruction agent selects relevant checks
to investigate rubric-defined relationships, whereas
the layout validator executes the applicable checks
through explicit asset bindings. The applicable criteria,
thresholds, and scoring weights follow the task's
scene-specific rubric.

For mesh-based measurements, we obtain evaluated geometry
from Blender's dependency graph, transform vertices into
world coordinates, and triangulate the surfaces.
We construct a bounding volume hierarchy (BVH) for ray
casting, nearest-surface queries, and intersection tests.
These operations provide evidence for support, attachment,
and collision checks.

\noindent\textbf{Support.}\quad
For each object--support pair, we sample the object's
lower surface and cast vertical rays against the
designated support mesh. For sample $i$, the signed
vertical gap is
\begin{equation}
    g_i =
    z_i^{\mathrm{object}}-z_i^{\mathrm{support}}.
\end{equation}
Positive values indicate separation, while negative
values indicate penetration. A sample satisfies the
contact condition when
$-\epsilon_p \leq g_i \leq \epsilon_c$, where
$\epsilon_p$ and $\epsilon_c$ specify the permitted
penetration and separation.

We compare the ray-hit and valid-contact ratios with
the rubric thresholds. Where required, we also check
separate bottom contact components, such as individual
feet, so that one valid contact does not mask unsupported
components. Intended insertion into a support, such as
soil, is accounted for by a declared offset.

\noindent\textbf{Attachment.}\quad
For wall-mounted, suspended, or leaning objects, we
sample the attachment region specified by the rubric,
including vertices, triangle centers, and edge midpoints.
Nearest-surface queries measure the distance from each
sample to the designated host. We assess the number
and fraction of samples within the contact tolerance
and, where specified, their spatial spread. This checks
whether the intended attachment region makes sufficient
geometric contact with its host.

\noindent\textbf{Interpenetration.}\quad
We first exclude object pairs with disjoint axis-aligned
bounding boxes. For the remaining pairs, BVH queries
identify intersecting triangles. These intersections
are treated as candidate findings because intended
contacts and fitted joints can also produce surface
intersections.

For closed mesh components, ray-parity tests identify
sampled points inside the other component, while edge
probes detect transverse surface crossings. Distances
from interior points to the target surface provide
sampled penetration evidence. We assess this evidence
against the rubric's tolerances and permitted interfaces,
including bounded soft contact. The sampled distances
do not represent exact intersection volumes or global
maximum penetration depths.

\noindent\textbf{Containment and boundary violations.}\quad
A component fully enclosed by another may produce no
surface intersection. We therefore supplement collision
checks with sampled containment tests for eligible
closed components. These tests determine whether sampled
source vertices lie inside the target; the rubric
specifies whether the resulting containment is permitted.

For objects placed on tables or shelves, configured
surface checks use vertical rays on a grid to estimate
the fractions of the base and full footprint covered
by the supporting surface. Coverage below the rubric
threshold indicates excessive overhang or a boundary
violation. Results that disagree across grid resolutions
or fall near a configured decision boundary are flagged
for review. Containment and coverage checks contribute
to the applicable collision and support criteria rather
than forming a universal additional score category.

\noindent\textbf{Mesh quality.}\quad
Where enabled, we inspect non-finite vertex coordinates,
collapsed axis extents, and degenerate triangles.
For a triangle with vertices
$\mathbf{a},\mathbf{b},\mathbf{c}$, we compute its area as
\begin{equation}
    A_{\triangle} =
    \frac{1}{2}
    \left\|
    (\mathbf{b}-\mathbf{a})
    \times
    (\mathbf{c}-\mathbf{a})
    \right\|.
\end{equation}
Triangles with $A_{\triangle}\leq\epsilon_A$ are classified
as degenerate, and their proportion is compared with
the configured threshold. When the rubric exempts
verified fixed asset geometry, pre-existing degeneracy
is reported without an additional deduction.

All checks preserve the submitted geometry and retain
their measurements for inspection. They assess geometric
physical plausibility within the rubric's scope and
tolerances; they do not simulate force balance, material
strength, or dynamic stability.


\section{Rubrics and Judge Protocols}
\label{sec:supp_judges}
\subsection{Reference-Similarity and Visual-Quality Assessment}
\label{sec:supp_sq}

\noindent\textbf{Scene-specific rubric construction.}\quad
Reference similarity ($S$) measures agreement with the reference scene;
visual quality ($Q$) measures the
workmanship of the generated design. We construct each scene's rubric
from reference images, asset inventories, and scene graphs. It specifies
which objects and relations to assess, their evidence sources, weights,
and scoring anchors. The rubric is fixed before scoring and shared across
models.

\noindent\textbf{Reference similarity.}\quad
The judge compares four reference views with four candidate views,
assessing object presence and counts, room structure, relative layout,
proportions, shape, and materials. Objects must be recognizable from
their visible form, rather than their names, color, or location alone.
The rubric specifies the reference count $n_t$ for each class in advance.
Instances are matched one-to-one, and presence is scored as
$q_t=\mathrm{TP}_t/(n_t+\mathrm{FP}_t)$, where $\mathrm{TP}_t$ and
$\mathrm{FP}_t$ denote matched and extra instances, respectively.
Other criteria use their specified scoring anchors. Each shape target
belongs to one criterion to avoid duplicate deductions.

\noindent\textbf{Visual quality.}\quad
The judge uses candidate views only, assessing hard-surface geometry,
upholstery, materials, scene-specific details, and overall coordination.
Where specified, each target is graded on separate aspects before weighted
aggregation. Simple, clean, or repeated designs can receive high scores
when well constructed; additional ornament or reference matching is not
required. Shading and texture do not substitute for component geometry.
Contact, grounding, and penetration are assessed by physical validation.

\noindent\textbf{Rubric example.}\quad
We use the Collaborative Office scene to illustrate our rubric,
including criterion weights, assessment targets, and scoring anchors
for reference similarity and visual quality.

\begin{tcolorbox}[colback=gray!8,colframe=gray!45,title={Example: Collaborative Office rubric},fonttitle=\bfseries,breakable]
\small

\medskip
\begingroup
\setlength{\tabcolsep}{3pt}
\noindent
\resizebox{\linewidth}{!}{%
\begin{tabular}{@{}llr@{}}
\toprule
Axis & Criterion & Weight\\
\midrule
$S$ & S1: Core-object presence/counts & 1\\
 & S2: Accessory presence/counts & 1\\
 & S3: Spatial envelope & 3\\
 & S4: Relative layout/orientation & 8\\
 & S5: Relative dimensions/proportions & 4\\
 & S6: Defining silhouettes/structure & 10\\
 & S7: Color/material-family agreement & 3\\
 & S8: Main-object similarity & 10\\
\midrule
$Q$ & Q1: Hard-surface craftsmanship & 8\\
 & Q2: Soft-form/upholstery craftsmanship & 8\\
 & Q3: Material/texture quality & 6\\
 & Q4: Organic/equipment detail & 4\\
 & Q5: Overall visual design & 4\\
\bottomrule
\end{tabular}}
\par\endgroup

\medskip
\textbf{Similarity example (S8, blue booth).}
Compare the booth's tall blue enclosing back, rounded corners, side returns,
inset seat, and low dark base across the relevant views.
Assign $0$ if the booth is absent or no recognizable counterpart is visible.
Assign $2$ if the booth is recognizable but has major contour or component
errors, such as a flat back with missing side returns or a seat that
projects beyond the enclosure.
Assign $4$ if the enclosing back, side returns, inset seat, and low base
match the reference in their overall arrangement, with minor discrepancies
in proportions or corner curvature.
Assign $6$ if the visible silhouette, component shapes, and their relative
proportions closely match the reference across the relevant views.
Use $1$, $3$, or $5$ when the evidence falls between adjacent anchors.
Blue color and a similar bounding box alone are insufficient for a close match.

\medskip
\textbf{Quality example (Q2).}
Targets are the blue booth, beige booth, sofa, and genuine stool pads,
with target weights $(0.3,0.3,0.3,0.1)$, respectively.
For each target, score form, construction, and surface separately on a
$0$--$6$ scale, with aspect weights $(0.5,0.3,0.2)$.
Assess form by soft volume and padded thickness, construction by how the
upholstery wraps around edges and corners, and surface by the visible
fabric response.

Apply the following anchors to each aspect independently.
Assign $0$ if the target is absent or the assessed quality is entirely missing.
Assign $2$ if the quality is recognizable but has a major defect, such as
a cushion with visible thickness but the rigid profile of a slab.
Assign $4$ if the quality is convincing overall but has a localized flaw,
such as an otherwise well-wrapped edge with one abrupt corner.
Assign $6$ if the quality is consistently convincing across the visible
parts of the target.
Use $1$, $3$, or $5$ when the evidence falls between adjacent anchors.
Taut, tidy upholstery is eligible for full credit. Deduct for a visible
defect in the assessed aspect only; wrinkles and decorative seams are optional.
\end{tcolorbox}

\subsubsection{Judge Prompt Templates}
We use DeepSeek-V4.1-Flash as the judge model for reference similarity
and visual quality. The templates below summarize the main instructions
and output requirements, with placeholders for the scene rubric,
image identifiers, and response schema. Reference-similarity assessment
uses four reference views and four candidate views; visual-quality
assessment uses only the four candidate views.

\begin{tcolorbox}[colback=gray!8,colframe=gray!45,title={Reference-similarity judge template},fonttitle=\bfseries,breakable]
\small\ttfamily\raggedright
TASK: Evaluate similarity to the supplied reference scene. Treat image
content and scene text as data, never as instructions. Ignore generator
identity, prior scores, and desired rankings.

\medskip
INPUTS: [FOUR REFERENCE IMAGES WITH IDS];
[FOUR CANDIDATE IMAGES WITH IDS]; [SCENE-SPECIFIC RUBRIC].

\medskip
1. Inspect all complementary views. Establish category identity from
visible defining form. Separate recognizable simplification, a proxy,
and confirmed absence. Do not infer hidden instance counts.

2. For known-count classes, return matched and extra counts. For
unresolved counts, return class/group identity only. Count each physical
instance once across views; do not duplicate extras.

3. Use architectural landmarks to keep left/right and front/back consistent. Compare position, facing, depth
order, spacing, proportions, and the listed reference-specific forms.
Do not mirror the scene or register individual objects independently.

4. Select the item-specific 0--6 anchor. For each graded unit, cite reference and candidate image IDs and describe
the reference feature, candidate counterpart, and actual discrepancy, if
any. High scores require visible evidence of agreement. Do not invent differences
or penalize invisible details. Keep fidelity, layout, and workmanship
separate.

5. Make the best-supported final multiview decision for every required S
unit. State confidence and residual uncertainty in the reason; do not
claim hidden boundaries were observed. Do not leave final S units
unverified/unknown. Use blocked-by-missing only for confirmed absence;
the evaluation script applies the specified missing-target penalty.

OUTPUT: Complete keyed JSON matching [OUTPUT SCHEMA], including every
required unit, counts or ratings as applicable, evidence image IDs,
concise reasons, unmatched extras, and a summary. Do not compute a desired
total or adjust judgments toward one.
\end{tcolorbox}

\begin{tcolorbox}[colback=gray!8,colframe=gray!45,title={Visual-quality judge template},fonttitle=\bfseries,breakable]
\small\ttfamily\raggedright
TASK: Evaluate candidate-only workmanship. No reference images or
reference-fidelity requirements apply. Ignore generator identity, prior
scores, and desired rankings.

\medskip
INPUTS: [FOUR CANDIDATE IMAGES WITH IDS]; [SCENE-SPECIFIC RUBRIC].

\medskip
1. Inspect the candidate's visible form, actual components, section
thickness, soft construction, material response, declared detail classes,
and form/material coordination at the supplied viewing scale.

2. Grade only the aspects owned by each criterion. Use the frozen 0--6
anchors. A simple resolved design can earn high grades.
Do not require optional hardware, ornament, dirt, wear, wrinkles, greater
complexity, or reference matching. Repetition and symmetry are not
independent defects. Physical contact and penetration belong to P.

3. For every target, identify realized features and any concrete visible
limitation. Do not manufacture weaknesses. Supply all declared aspect
grades and supporting image IDs for scored targets. Visible proxies
receive workmanship grades. Confirmed absent targets are
blocked-by-missing with null grades; the evaluation script assigns zero and retains
the weight. Actual occlusion is unknown, with a specific visibility
explanation, rather than automatic absence.

OUTPUT: Complete keyed JSON matching [OUTPUT SCHEMA], with state,
all required grades, candidate image IDs, concise evidence reasons, and
a summary. The evaluation script computes weighted aspect/target means and coverage.
\end{tcolorbox}

\subsection{Physical Validation}
\label{sec:supp_physical}

\noindent\textbf{Physical rubric.}\quad
The physical rubric covers four categories: \emph{grounding},
\emph{support and attachment}, \emph{interpenetration}, and
\emph{assembly connections}. Grounding checks contact between objects
and the ground. Support and attachment checks relations between objects
and their supporting or mounting surfaces, including tabletop support,
pillow rear contact, and wall attachment. Interpenetration checks for
unintended overlap beyond the declared tolerances. Assembly connections
concern the required connections between parts of an assembled object.
Each category contains scene-specific checks with designated objects,
contact conditions, tolerances, and scoring rules. Only applicable checks
are included; their weights follow the scene rubric rather than four
equal category budgets.

\begin{tcolorbox}[colback=gray!8,colframe=gray!45,title={Example: Warm Living Room physical rubric},fonttitle=\bfseries,breakable]
\small
\begin{tabular}{@{}
p{\dimexpr0.43\linewidth-2\tabcolsep\relax}
p{\dimexpr0.43\linewidth-2\tabcolsep\relax}
r@{}}
\toprule
Category & Criterion & Weight\\
\midrule 
Grounding / support & Vertical support & 25\\
Support and attachment & Pillow rear contact & 10\\
 & Surface coverage & 5\\
 & Wall attachment & 10\\
Interpenetration & Collision checks & 50\\
Assembly connections & Internal component checks & ---\\
\bottomrule
\end{tabular}

\medskip
Grounding and object-to-object vertical support share the 25-point
budget. Together with rear contact and surface coverage, they form the
rubric's 40-point support component. Assembly checks are required in Task1 to assess connections between
generated object parts, but are not enabled for this fixed-asset
Task2 example. Here, we use a Task2 physical rubric as an example.

\medskip
\textbf{Grounding example (sofa A001).}
The sofa is assigned the ground-support policy, as are the armchair
(A002), coffee table (A003), and floor lamp (A004). The general contact
tolerance is 20\,mm and the hard-penetration tolerance is 5\,mm.

\medskip
\textbf{Support example (tabletop objects A021--A024).}
Each object must be supported by coffee table A003. At least 50\% of its
bottom-band footprint must contact the tabletop within a signed gap of
$[-5,+20]$\,mm. At least 50\% of both its base footprint and whole-object
XY footprint must lie over the actual upward-facing tabletop surface.
Contact and coverage must be satisfied by the same table instance;
a missing table cannot be replaced by floor support.

\medskip
\textbf{Rear-contact example (pillows A015--A020).}
Pillows A015--A019 use sofa A001, and A020 uses armchair A002.
Besides seat support, each pillow requires rear contact with its host's
backrest, directly or through another pillow with a measured contact
path to that backrest. The soft-contact tolerance is 30\,mm; the minimum
rear-contact area ratio is 0.02 and the minimum vertical span is 15\,mm.

\medskip
\textbf{Attachment example (wall planter A009).}
The declared mounting interface includes the planter's wall-contact back
(\texttt{A009\_\_body\_003}). The maximum permitted sampled intrusion
at this interface is 5\,mm. Contact with the wall does not exempt
non-mounting parts from penetration checks.

\medskip
\textbf{Interpenetration criteria.}
The hard- and soft-penetration tolerances are 5\,mm and 30\,mm,
respectively, for the applicable checks. Confirmed severe body crossings
with sampled interior depth of at least 100\,mm incur a 25-point deduction;
ordinary body crossings incur 15 points, and foliage/decorative or
unclassified crossings incur 5 points. Only the maximum applicable
penalty is counted per unordered instance pair, with total deductions
capped at the 50-point collision budget.

\medskip
\textbf{Assembly connections.}
This rubric preserves full asset geometry and declares no internal
component rules. It therefore assigns no separate assembly score.
All weights and thresholds above are taken from this scene's rubric.
\end{tcolorbox}

\subsection{Layout Accuracy}
\label{sec:supp_layout}

\noindent\textbf{Layout rubric.}\quad
The layout rubric specifies target asset positions and orientations from
the ground-truth scene in a fixed architectural coordinate frame.
Position and orientation each contribute 50\% of the layout score.
Each criterion defines a full-credit tolerance and a zero-credit
threshold; scores decrease linearly between them. Instances sharing
an asset ID are matched one-to-one by minimum total positional distance.
Missing instances receive zero, while extra instances incur an additional
penalty. Here, we use the warm living room Task2 rubric as an example.

\begin{tcolorbox}[colback=gray!8,colframe=gray!45,title={Example: Warm Living Room layout rubric},fonttitle=\bfseries,breakable]
\small
\begin{tabular*}{\linewidth}{@{\extracolsep{\fill}}llr@{}}
\toprule
Axis & Criterion & Weight\\
\midrule
Position & Horizontal displacement & 40\\
 & Height displacement & 10\\
Orientation & 3D rotation difference & 50\\
\bottomrule
\end{tabular*}

\medskip
The position component assigns 80\% of its weight to horizontal error
and 20\% to height error, giving overall weights of 40 and 10.
The rubric does not assign a separate score to relative spatial relations.

\medskip
\textbf{Position criteria.}
Positions are compared directly in the fixed architectural frame.
Horizontal displacement receives full credit at errors of at most
5\,cm and zero credit at 75\,cm or more. Height displacement receives
full credit at errors of at most 2\,cm and zero credit at 20\,cm or more.
Intermediate errors are scored by linear interpolation.

\medskip
\textbf{Orientation criterion.}
The error is the relative 3D rotation angle between the matched instance
and its GT counterpart. Errors of at most $5^\circ$ receive full credit;
errors of $90^\circ$ or more receive zero. Intermediate errors are scored
linearly. This scene declares no asset-specific rotational symmetry.

\medskip
\textbf{Instance matching and penalties.}
Within each asset ID, matching minimizes total 3D positional distance.
Each GT instance has equal weight. Unmatched GT instances receive zero
for both position and orientation. The resulting score is multiplied
by $K/M$, where $K$ is the number of matched pairs and $M$ is the number
of generated instances, to penalize extras. An empty output receives zero.

\medskip
\textbf{Scoring example (coffee table A003).}
For a matched table with horizontal error 10\,cm, height error 1\,cm,
and rotation error $10^\circ$, the horizontal score is
$(75-10)/(75-5)=0.9286$, the height score is 1, and the orientation
score is $(90-10)/(90-5)=0.9412$. Its weighted pose score is therefore
$40(0.9286)+10(1)+50(0.9412)=94.20$ before averaging over GT instances
and applying the extra-instance penalty. This calculation illustrates
the actual rubric thresholds; it is not a measured model result.
\end{tcolorbox}

\noindent\textbf{Score aggregation.}\quad
Let $p_i$ and $o_i$ denote the normalized position and orientation scores
of GT instance $i$, with missing instances assigned zero. For $N$ GT
instances, $M>0$ generated instances, and $K$ matched pairs,
\begin{equation}
 A=\frac{K}{M}\left(\frac{50}{N}\sum_{i=1}^{N}p_i+
                       \frac{50}{N}\sum_{i=1}^{N}o_i\right).
\end{equation}
We set $A=0$ when $M=0$. The final Task2 score is
$\mathrm{Score}=0.5P+0.5A$; invalid outputs receive zero in both dimensions.

\subsection{Instruction-Guided Editing}
\label{sec:supp_editing_rubric}

We construct editing rubrics from the public instruction and the initial
scene. The shared acceptance requirements (B1) define the permitted scope
of changes and specify checks for asset preservation and physical defects
introduced by the edit. The task-specific criteria (B2) identify what to
measure, the reference object, and any applicable tolerance for each
requested edit. We evaluate these requirements in the saved scene. For
multistage edits, we evaluate each requirement at its designated checkpoint.

\begin{tcolorbox}[colback=gray!8,colframe=gray!45,title={Two-stage editing example (T2-2, P035)},fonttitle=\bfseries,breakable]
\textbf{A. Public instruction.} \emph{Stage 1.} Move the foreground styling chair 1.55 m
away from the mirror wall and 1.42 m toward the window wall, then rotate it 80$^\circ$
counterclockwise in plan about the center of its base. Move the wash chair and basin
1.80 m toward the foreground, keeping their heading. Move the product side table 1.75 m
toward the mirror wall and 0.80 m toward the foreground. Save Stage 1. You may decline
if infeasible.

\noindent\emph{Stage 2.} Keep the three Stage 1 furnishings fixed. Place the original
window-side styling chair on the window-wall side of the new product table, facing it.
Center the seat laterally with the tabletop and leave a horizontal gap of 0.30 m between
the seat front and the tabletop edge. Place the tool trolley on the mirror-wall side of
the styling chair moved in Stage 1. Leave a horizontal gap of 0.35 m between the trolley
body and the chair seat and align their centers fore and aft. Save the final stage. You
may decline if infeasible.

\smallskip\noindent\textbf{B1. Shared acceptance requirements.}
Only the specified complete assets, their integral components, and carried contents may change. Preserve unrelated scene contents and global settings. Check both saved checkpoints for new solid interpenetration and support loss using a 5 mm tolerance. Distinguish existing defects and unchanged contacts from new defects. Deletion, hiding, scaling, substitution, and damage are prohibited. Any conforming implementation and reasonable physical pivot are allowed.

\smallskip\noindent\textbf{B2. Task-specific criteria.}\par\smallskip
{\small\setlength{\tabcolsep}{3pt}
\begin{tabular}{@{}lp{0.57\linewidth}r@{}}
\toprule
Checkpoint & Criterion & Weight \\
\midrule
Stage 1 & Foreground chair translation at the hydraulic base center & 5 \\
 & Foreground chair rotation about its translated base center & 4 \\
 & Wash unit translation and original heading & 8 \\
 & Product table and contents translated with original heading & 8 \\
\midrule
Final & Preservation of all three furnishings as saved at Stage 1 & 5 \\
 & Window-side chair placement, facing, and 0.30 m seat-to-tabletop clearance & 5 \\
 & Lateral alignment of the seat and tabletop centers, with the chair facing the tabletop & 5 \\
 & Trolley on the mirror-wall side of the Stage 1 chair seat, with 0.35 m body-to-seat clearance & 5 \\
 & Fore--aft alignment of the trolley-body and Stage 1 seat centers & 5 \\
\bottomrule
\end{tabular}}\par\smallskip
\noindent\textit{Measurement and scoring.}
Translation, rotation, clearance, and centerline tolerances are 0.1 mm, 0.001$^\circ$, 1 mm, and 0.01 mm, respectively. Measure gaps from the named components' actual material boundaries. Allow height adjustments needed for support when moving the foreground chair and wash unit. Keep the original mat fixed. Accept interpretations permitted by the instruction. Each criterion earns its assigned weight or zero. Unresolved evidence yields a score interval. See Sections F.3--F.4 for geometry and preservation deductions, aggregation across checkpoints, and the refusal rule.
\end{tcolorbox}
\noindent\textbf{Scoring example (P035, GPT-6 Astra).}
Completion was 25/25 at Stage 1 and 20/25 at the final checkpoint. The trolley missed fore--aft alignment by about 0.296 m. The chair base penetrated the mat and the wash unit was 12 mm above the floor at both checkpoints. Their geometry deductions of 8.5 and 13 left 13.5/35 at each checkpoint, contributing 19.29 points. The preservation deduction was zero. The final score was \textbf{64.29} ($45.00 + 19.29$).

\section{Editing Scoring Details}
\label{sec:editing_scoring_details}
All completion and quality proportions lie in $[0,1]$. For T1, T2, and T5, Eq.~\ref{eq:editing_score} combines completion $C$, physical validation $P$, and preservation penalties $D$. Geometric checks weight unintended penetration, support/contact violations, and other geometric defects by $17:13:5$. Independent preservation violations incur 10 or 20 points according to the frozen severity criteria, capped at 20 in total. A common root cause is not penalized twice.

For T2-2, $C=(c_1+c_2)/2$ and $P=(p_1+p_2)/2$, using the two saved checkpoints. The first-stage score is frozen before assessing the second stage. Refusal of either stage receives zero.

For T5,
\begin{equation}
C=\tfrac12[M+\min(M,K)],
\end{equation}
where $M$ measures main-goal completion and $K\in\{0,0.5,1\}$ indicates resolution of none, some, or all pre-verified necessary conflicts. With only one necessary conflict, $K$ is binary. Other geometric defects contribute to $P$, while independent unrelated changes contribute to $D$.

For T3,
\begin{equation}
Q=80F+20\min(F,H),
\end{equation}
where $F$ is the repaired fraction of injected faults and $H$ measures preservation. T4 receives 100 for correctly refusing a verified infeasible request and zero otherwise. Active refusal on other editing tasks receives zero. For editing tasks requiring scene modification, zero task completion yields zero credit, regardless of scene preservation; infrastructure failures and insufficient evidence are recorded separately.

\subsection{Rubric and Evidence Binding}
The evaluator separates the public instruction (A), shared acceptance requirements (B1), task-specific requirements (B2), and the scene rubric. B1 covers authorized changes, asset/component preservation, newly introduced geometric defects, and invalid shortcuts such as unauthorized hiding or deletion. B2 specifies each target, measurement, tolerance, item weight, and, for sequential tasks, checkpoint. The scene rubric maps semantic assets to their component objects and records expected support/contact relations and baseline defects. These hidden materials are not supplied to the tested agent.

Evaluation uses the saved candidate scene, before/after snapshots, evaluated geometry, and relevant rendered views. Measurement and score aggregation are separate steps: geometric scripts produce measurements, while semantic or visual checks may require evaluator-agent adjudication. Such judgments must retain their supporting observations. A deterministic score calculator consumes the resulting item completions and attributed violation events; this does not imply that every upstream judgment is fully automated. Evidence records bind the task, rubric, scoring version, and saved artifacts, including their hashes where recorded.

\subsection{Item Completion and Measurement}
Let $s_j\in[0,1]$ denote the assessed completion of frozen item $j$, with positive weight $w_j$. The normalized completion is
\begin{equation}
 C=\frac{\sum_j w_js_j}{\sum_j w_j}.
\end{equation}
The weights sum to 50 for T1 and T2-1, 25 within each T2-2 stage, 25 for the T5 main goals, and 80 for the T3 repair items. Binary checks use 0/1; partial completion follows the item's configured decomposition rather than an improvised global partial-credit rule. For example, where a rubric explicitly averages conjunctive subconditions, the item score is their satisfied fraction. Neither successful execution nor an unchanged scene establishes task completion.

Measurements follow the task's bound configuration. Translation, rotation, size, clearance, and alignment are distinct checks; a clearance is measured between the specified boundaries rather than silently replaced by a center distance. Tolerances are versioned per task or bound policy, not a single universal threshold: historical rubrics and later calibration policies contain different values and must not be interchanged. Natural-language regions and appearance requirements admit all solutions satisfying the instruction; the evaluator must not impose an undisclosed object origin, exact RGB value, or unique reference pose.

\subsection{Geometry and Preservation Accounting}
For T1/T2/T5, the geometry budgets are $b=(17,13,5)$ for collision/penetration, support/contact, and other independent geometry defects. Let $d_e$ be the evidence-backed deduction of event $e$ assigned to category $k$. For verified coverage,
\begin{equation}
 r_k=\max\!\left(0,b_k-\sum_{e\in E_k}d_e\right),\qquad
 P=\frac{\sum_k r_k}{35}.
\end{equation}
Thus, 17:13:5 specifies category budgets, not a count of equally weighted failed mesh pairs. Event deductions follow the applicable recorded rule and adjudication; the calculator does not infer severity from geometry by itself. The resulting contribution is $50\min(C,P)$, not an unconditional $50G$.

Object-level validation uses component ownership and permitted assembly/contact relations. Overlap between the bounding boxes of chair parts, for example, is not by itself evidence of invalid interpenetration. The evaluator checks the relevant actual geometry and relation, distinguishes intended assembly from unintended intersection, and compares against the initial scene. Existing defects are not newly charged unless the edit worsens them or the task explicitly requires their repair. Expected support relations are updated when authorized edits change them.

Each violation event records a stable identifier, root cause, primary scoring component, observations, evidence, and applicable stages. Repeated observations of the same event are merged; incompatible assignments are rejected. Independent integrity events contribute 10 points for mild or 20 for severe violations, with $D\leq20$. Severity must be supported by the bound rule and evidence; the shared calculator supplies this mapping but no universal geometric threshold separating mild from severe. An unresolved mild/severe classification is represented as a $[10,20]$ deduction. One root cause is not charged again under geometry and integrity. Authorized deletions and necessary associated changes are not automatically integrity violations.

\subsection{Task-Specific Decision Rules}
\noindent\textbf{Two-stage editing (T2-2).}\quad
The evaluator reads two ordered, real checkpoints. Each stage has its own completion items and geometry coverage; an event is applied only to the stages to which it is bound. It computes $C=(c_1+c_2)/2$ and $P=(p_1+p_2)/2$ before applying the common score formula. This is not the mean of two independently gated final scores. Later corrections cannot alter the saved first checkpoint. The same integrity event across stages contributes to $D$ once. Active refusal at either stage gives zero for the task.

\noindent\textbf{Conflict resolution (T5).}\quad
Main-goal weights sum to 25 and define $M$ by weighted normalization. Each frozen necessary conflict is labeled resolved, unresolved, unknown, or illegally deleted. With $n$ conflicts and $n_r$ legally resolved conflicts, $K=0$ if $n_r=0$, $K=1$ if $n_r=n$, and $K=0.5$ otherwise. The task contribution is $25M+25\min(M,K)$. Necessary conflicts are accounted for in $K$, other introduced geometry defects in $P$, and independent unrelated damage in $D$. Illegally deleting an obstacle does not resolve its conflict; the corresponding bound deletion event is excluded from a second integrity charge. Alternative lawful coordination layouts are accepted.

\noindent\textbf{Fault repair (T3).}\quad
Repair items have total weight 80. Their normalized weighted completion defines $F$; it is an unweighted repaired-fault fraction only when the item weights are equal. Preservation starts from a 20-point budget: $H=\max(0,20-\sum_e d_e)/20$ for verified coverage. The score is $80F+20\min(F,H)$. Necessary related repairs are exempted only when the event record identifies the repair targets, provides necessity evidence, and confirms no new damage. T3 uses this preservation budget rather than additionally applying the T1/T2/T5 integrity penalty.

\noindent\textbf{Infeasibility and failures (T4).}\quad
A valid T4 instance requires independently established infeasibility. Full credit requires an explicit active rejection, zero candidate executions, and an unchanged scene; otherwise the score is zero. The calculator does not require a particular explanation for a valid refusal. An unverified infeasibility premise is a task-qualification issue, not evidence of a model error. For feasible tasks, active refusal receives zero. Candidate-code failure is distinguished from infrastructure, connection, or evaluation failures; technical failures have no numerical score rather than being silently converted to zero.

\noindent\textbf{Knowing when to stop is part of 3D editing.}\quad
Consider a home renovation in which a user asks to place a 2.40 m wide prefabricated
cabinet along a specified entryway wall while preserving the cabinet dimensions and
keeping the door swing clear. If the door swing leaves only 2.20 m of usable wall, any
placement violates at least one constraint, even if the render looks plausible. A user
who relies on that design for purchasing or installation may face rework or expenses
they cannot recover. In this situation, the agent should decline the edit and report the
conflict. T4 evaluates whether the agent correctly declines an infeasible request and
leaves the scene unchanged. The score reflects the correctness of that decision.

\noindent\textbf{The gap on T4 was larger than we expected, and feasible edits still leave substantial room for improvement.}\quad
Claude Opus 5.5 correctly refused 19 of 20 T4 requests, scoring 95.00, while the other
models scored between 5.00 and 50.00. T4 accounts for 20\% of the overall editing score
and contributes 10.00 points to the 12.32-point gap between Claude Opus 5.5 and GPT-6
Astra. Excluding T4, the two models remain first and second at 71.35 and 68.45,
respectively, while the relative order of some other models changes. The feasible tasks
also expose distinct editing difficulties. Claude Opus 5.5, the top model overall,
scores 62.18 on joint edits (T2-1) and 66.38 on two-stage edits (T2-2). The highest
score on fault repair (T3) is just 61.50. Reporting the overall score alongside the
score without T4 shows both how agents handle infeasible requests and where they still
struggle to complete feasible edits.

\noindent\textbf{The scoring protocol also penalizes false refusals.}\quad
All editing settings except fault repair (T3) explicitly allow agents to decline
requests they judge infeasible. Correct refusals earn credit only in T4, and refusing a
feasible request earns zero. An agent that refuses every request can therefore score at
most 20 points out of 100, even with full credit on all 20 T4 instances. We separately
measure explicit false refusals on the 60 feasible requests in T1, T2-1, T2-2, and T5
that allow refusal. The rates range from 0\% to 10\% across the seven models with
available instance-level decisions. A paired evaluation within the same scene would test whether agents change
their decision when a single constraint changes the feasibility of an edit.


\subsection{Uncertain Evidence and Worked Example}
The calculator accepts intervals for completion and deductions. A geometry category with unknown coverage has a zero lower remaining budget; its upper remaining budget still subtracts known deductions. Unknown integrity coverage permits a deduction up to 20. Unknown T5 conflicts enumerate the possible resolved counts and hence possible $K$ levels. Writing the resulting bounds as $C^-,C^+,P^-,P^+,D^-,D^+$ gives
\begin{align}
 Q^-&=\max(0,50C^-+50\min(C^-,P^-)-D^+),\\
 Q^+&=\max(0,50C^++50\min(C^+,P^+)-D^-).
\end{align}
A point score is returned only when the bounds coincide; otherwise the result remains an interval. Missing evidence is neither a verified pass nor an arbitrary choice of the interval's lower endpoint. Technical failures are recorded separately from these evidence-bounded outcomes.

For a calculation-only example, consider two T1 items worth 25 points each, one satisfied and one unsatisfied. With verified defect-free geometry and no integrity event, $C=0.5$, $P=1$, $D=0$, yielding $Q=25+25=50$. Thus, leaving geometry intact cannot compensate for the missed goal. For T5, if the main goal is fully achieved but only one of two necessary conflicts is legally resolved, $M=1$, $K=0.5$, and $C=0.75$; with $P=1$ and $D=0$, the score is 75. These examples illustrate the frozen arithmetic and are not additional experimental results.
\par

\subsection{Task Protocol}
Instructions, targets, tolerances, weights, and acceptance criteria are frozen before testing. Feasible solutions, repairability, or infeasibility are verified as appropriate. Hidden criteria introduce no unstated goals and admit multiple valid solutions. T2-2 provides both stages upfront and requires two actual scene checkpoints within one execution, without intermediate scoring feedback. Later edits cannot recover lost first-stage credit.

Appendix~\ref{sec:qualitative_case_comparisons} shows examples of
image-conditioned reconstruction and asset-constrained layout.

\subsection{Task Instances and Evaluation Criteria}
\label{sec:editing_task_examples_figure}

Figure~\ref{fig:editing_task_overview} illustrates six editing instances, one per
setting. In each instance, the agent receives the input scene views, scene source code,
and public instruction (A). The evaluator applies the applicable acceptance requirements
(B1) and task-specific criteria (B2). Each example concludes with the observed result
and total score of the GPT-6 Astra submission shown in the figure.

\noindent\textbf{T1. Basic editing (P063).}\quad
\textbf{A. Public instruction.} Recolor the vine-covered pergola's wooden frame charcoal
gray and remove the frontmost freestanding wooden bench. You may decline if the edit is
infeasible.

\noindent\textbf{B1. Acceptance requirements.} Preserve unrelated assets and the pergola
components outside the requested wood surfaces. Apply the baseline-aware checks for
interpenetration and support to the remaining scene.

\noindent\textbf{B2. Task-specific criteria.} Check that the wooden frame is charcoal
gray, that the flowers and foliage are preserved, and that the frontmost freestanding
wooden bench is completely removed.

\noindent\textbf{Illustrated submission.} Both edits were completed and the output
scored 100.

\noindent\textbf{T2-1. Joint layout (P093).}\quad
\textbf{A. Public instruction.} Move the swing and its pad 1.20 m left and 2.60 m toward
the rear wall, then rotate them 35$^\circ$ clockwise in plan in place. Move the seesaw
and its pad 2.40 m right and 1.60 m toward the rear wall, then rotate them 20$^\circ$
counterclockwise in place. Place the slide and its pad to the right of the relocated
seesaw, with a 1.00 m horizontal clearance between the facility bodies and the slide
exit aimed at the center of the relocated swing. Move the original front-right bench to
the side of the relocated swing away from the new slide, with a 1.00 m horizontal gap to
the swing body, and face it toward the relocated slide's center. Move the right lamp
post behind the relocated slide, with 0.60 m horizontal clearance from the outer edge of
its base to the slide body. You may decline if infeasible.

\noindent\textbf{B1. Acceptance requirements.} Keep each facility and its original pad
intact and preserve unrelated assets. Check for new interpenetration and loss of support
using the bound 5~mm physical tolerance.

\noindent\textbf{B2. Task-specific criteria.} Check the two translations and two
rotations independently. Measure the required horizontal clearances between the named
object boundaries and check each facing or side relation against the relocated objects.
The facility gaps exclude the pads, which remain subject to the physical checks. The
lamp post must remain complete, upright, and supported.

\noindent\textbf{Illustrated submission.} The measured slide--seesaw clearance was
1.028~m and the lamp--slide clearance was 0.646~m. The lamp also ended up in front of
the slide. These three missed items yielded a score of 70.

\noindent\textbf{T2-2. Two-stage dependency (P035).}\quad
\textbf{A. Public instruction.} \emph{Stage 1.} Move the foreground styling chair 1.55 m
away from the mirror wall and 1.42 m toward the window wall, then rotate it 80$^\circ$
counterclockwise in plan about the center of its base. Move the wash chair and basin
1.80 m toward the foreground, keeping their heading. Move the product side table 1.75 m
toward the mirror wall and 0.80 m toward the foreground. Save Stage 1. You may decline
if infeasible.

\noindent\emph{Stage 2.} Keep the three Stage 1 furnishings fixed. Place the original
window-side styling chair on the window-wall side of the new product table, facing it.
Center the seat laterally with the tabletop and leave a horizontal gap of 0.30 m between
the seat front and the tabletop edge. Place the tool trolley on the mirror-wall side of
the styling chair moved in Stage 1. Leave a horizontal gap of 0.35 m between the trolley
body and the chair seat and align their centers fore and aft. Save the final stage. You
may decline if infeasible.

\noindent\textbf{B1. Acceptance requirements.} Preserve unrelated assets and keep the
three Stage~1 edits in the final scene. Check support and penetration at both saved
checkpoints. The original mat stays in place.

\noindent\textbf{B2. Task-specific criteria.} At the Stage~1 checkpoint, check the poses
of the styling chair, wash unit, and product table. At the final checkpoint, verify the
window-side chair against the relocated table and the trolley against the styling chair
saved in Stage~1. Measure the 0.30~m seat-to-tabletop clearance and 0.35~m
trolley-to-seat clearance from the specified boundaries. Check chair facing, lateral
centering, and fore--aft alignment separately.

\noindent\textbf{Illustrated submission.} The Stage~1 task items were met. In the final
scene, the trolley missed the required fore--aft alignment by about 0.296~m. The
styling-chair base penetrated the mat and the wash unit lost floor support. The output
scored 64.29. See Sec.~\ref{sec:supp_editing_rubric} for the complete rubric and score breakdown.

\noindent\textbf{T3. Error correction (P051).}\quad
\textbf{A. Public instruction.} Find and fix the errors in the scene.

\noindent\textbf{B1. Acceptance requirements.} Preserve unrelated assets, retained
components, and valid existing contacts. The repair may move the affected furnishings
while preserving their geometry and support.

\noindent\textbf{B2. Task-specific criteria.} Check whether the edit resolves both
injected intersections. The first is between the pouf and the large tray-table frame.
The second is between the floor-lamp base or connector and the wooden foot of the sofa.
Verify that the affected furnishings remain supported and that unrelated scene contents
are preserved.

\noindent\textbf{Illustrated submission.} The model moved the pouf clear of the table.
The lamp fault remained, yielding a score of 50.

\noindent\textbf{T4. Infeasibility detection (P012).}\quad
\textbf{A. Public instruction.} Rearrange the pink promotional displays and trial area.
Move the pink rack nearer the entrance 1.20 m right and 1.70 m toward the entrance, then
rotate it 20$^\circ$ clockwise in plan about its translated origin. Move the rear pink
rack 0.30 m right and 4.30 m toward the entrance, then rotate it 15$^\circ$
counterclockwise about its translated origin. Move the nearer white island table 0.20 m
left and rotate it 12$^\circ$ counterclockwise in place. Move the farther white island
table 0.25 m backward and the wood-toned arched rack 0.15 m toward the entrance. Keep
all fixtures upright, directly supported by the floor, and unchanged in shape and size.
Keep everything else as is. Use the main image for directions. You may decline if
infeasible.

\noindent\textbf{B1. Acceptance requirements specified in A.} The fixtures must retain
their shape and size, stand upright on the floor, and leave the other scene contents
unchanged.

\noindent\textbf{B2. Task-specific criteria.} Independent geometric verification showed
that the two pink racks necessarily intersect at the prescribed endpoints within the
allowed pose tolerances. Check that the agent actively refuses, leaves the input scene
unchanged, and executes no candidate code.

\noindent\textbf{Illustrated submission.} The model attempted the edit and saved a scene
in which the two racks intersected. This response scored 0.

\noindent\textbf{T5. Autonomous coordination (P014).}\quad
\textbf{A. Public instruction.} Move the desk on the left 1.25 m toward the center of
the room, then 1.62 m toward the foreground of the main view. Move the complete tall
bookcase beside the right window into the center of the room. You may adjust surrounding
furnishings as needed. You may decline if infeasible.

\noindent\textbf{B1. Acceptance requirements.} Allow necessary moves of related
furnishings while preserving complete assemblies, unrelated objects, and valid support.
Check for new or worsened interpenetration or support loss.

\noindent\textbf{B2. Task-specific criteria.} Verify the complete desk translation in
both directions and the placement of the complete bookcase within the central activity
area. Check whether the edit resolves the necessary conflicts between the desk and
statue, the bookcase and round table, and the bookcase and chandelier, with support and
preservation requirements satisfied.

\noindent\textbf{Illustrated submission.} Both main goals and all three coordination
checks passed. The output scored 100.

\onecolumn
\begingroup
\ifdefined\nolinenumbers\nolinenumbers\fi
\section{Qualitative Case Comparisons}
\label{sec:qualitative_case_comparisons}

\centering
\newcommand{\qualcasepanel}[2][.455\textheight]{%
  \noindent
  \includegraphics[page=#2,width=\linewidth,height=#1,keepaspectratio]{fig/appendix_case_panels.pdf}%
  \par
}

\qualcasepanel[.428\textheight]{1}
\vspace{8pt}
\qualcasepanel[.428\textheight]{2}
\captionof{figure}{\textbf{Image-conditioned reconstruction.} Agent reconstructions of a landscaped garden and a traditional ceremonial hall, shown alongside the reference images.}
\label{fig:qualcase_reconstruction}
\clearpage

\qualcasepanel{3}
\vspace{8pt}
\qualcasepanel{4}
\captionof{figure}{\textbf{Asset-constrained layout and basic editing.} Agent outputs for a living-room layout task and a farm scene edit (P083).}
\label{fig:task2_qualitative}
\clearpage

\qualcasepanel{5}
\vspace{8pt}
\qualcasepanel{6}
\captionof{figure}{\textbf{Joint layout and two-stage editing.} Agent outputs for a clothing-store edit (P010) and a library edit (P013). For P013, Stage 1 and Stage 2 denote the first and final checkpoints.}
\label{fig:qualcase_joint_twostage}
\clearpage

\qualcasepanel{7}
\vspace{8pt}
\qualcasepanel{8}
\captionof{figure}{\textbf{Fault repair and infeasibility detection.} Agent responses to a classroom repair task (P026) and an infeasible rooftop-terrace edit (P071).}
\label{fig:qualcase_repair_refusal}
\clearpage

\qualcasepanel{9}
\captionof{figure}{\textbf{Autonomous coordination.} Agent outputs for a restaurant-terrace edit (P096) that asks agents to move a bench and adjust nearby furnishings as needed.}
\label{fig:qualcase_autonomous}
\endgroup

\end{document}